\documentclass[10pt,journal,compsoc,hidelinks]{IEEEtran}
\usepackage{cite}

\ifCLASSINFOpdf
\else
\fi
\ifCLASSOPTIONcompsoc
 \usepackage[caption=false,font=normalsize,labelfont=sf,textfont=sf]{subfig}
\else
 \usepackage[caption=false,font=footnotesize]{subfig}
\fi
\usepackage[acronym,nonumberlist,toc]{glossaries}

\usepackage{graphicx}

\usepackage{amsmath}
\usepackage{amssymb}
\usepackage{todonotes}

\usepackage{booktabs}
\usepackage{colortbl}
\usepackage{multirow}
\usepackage{tabularx}
\usepackage{xcolor}
\usepackage{bm}

\usepackage[hidelinks]{hyperref}

\newcommand{\musub}[1]{\bm{\mu}_{#1}}
\newcommand{\Msub}[1]{\bm{M}_{#1}}
\newcommand{\Sigmasub}[1]{\bm{\Sigma}_{#1}}
\newcommand{\cn}[1]{c_{#1}}
\newcommand{\hadamard}{\circ}

\newcommand{\musubbf}[1]{\bm{\mu}_{\bm{#1}}}
\newcommand{\Sigmasubbf}[1]{\bm{\Sigma}_{\bm{#1}}}

\newcommand{\e}[1]{\mathbb{E}[#1]}
\newcommand{\ej}[2]{\mathbb{E}_{-#1}[#2]}
\newcommand{\elog}[1]{\mathbb{E}[\log #1]}
\newcommand{\elogj}[2]{\mathbb{E}_{-#1}[\log #2]}

\newcommand{\transpose}[1]{{#1}^{\top}}
\newcommand{\inverse}[1]{#1^{-1}}
\newcommand{\Identity}[1]{\textbf{I}_{#1}}

\newcommand{\entropy}[1]{\text{h}(#1)}
\newcommand{\diag}[1]{\text{diag}(#1)}
\newcommand{\trace}[1]{\text{Tr}(#1)}

\newcommand{\tensor}[1]{}

\newcommand\tran{\mkern-3mu{}_{}^{\scriptstyle\top}\mkern-4mu}

\newcommand{\ncomps}{M}
\newcommand{\ncompslc}{m}

\newcommand{\nslabslc}{k}

\newcommand{\fullx}{\mathcal{X}}

\newcommand{\fullp}{\mathcal{P}}

\newcommand{\varx}{\bm{X}_\nslabslc}
\newcommand{\vara}{\bm{A}}
\newcommand{\varc}{\bm{C}}
\newcommand{\vard}{\bm{D}_\nslabslc}
\newcommand{\varf}{\bm{F}}
\newcommand{\varp}{\bm{P}_\nslabslc}
\newcommand{\vare}{\bm{E}_\nslabslc}

\renewcommand{\varsigma}{\bm{\tau}}
\newcommand{\varalpha}{\bm{\alpha}}

\newcommand{\varsigmaa}{a_{\varsigma}}
\newcommand{\varsigmab}{b_{\varsigma}}

\newcommand{\varcele}{c_{\nslabslc\ncompslc}}
\newcommand{\varaele}{a_{i\ncompslc}}

\newcommand{\varxvec}{\bm{x}_{i\cdot \nslabslc}}
\newcommand{\varavec}{\bm{a}_{i\cdot}}
\newcommand{\varacol}{\mathbf{a}_{\cdot i}}
\newcommand{\varcvec}{\bm{c}_{\nslabslc\cdot}}
\newcommand{\vardvec}{\bm{D}_\nslabslc}
\newcommand{\varfvec}{\bm{f}_{\ncompslc\cdot}}

\newcommand{\varsigmavec}{\tau_\nslabslc}
\newcommand{\varalphavec}{\alpha_\ncompslc}

\newcommand{\varparameters}{\bm{\theta}}
\newcommand{\varfactorj}{\theta_j}
\newcommand{\allparameters}{\vara,\varc,\varf,\fullp,\varsigma,\varalpha}

\newcommand{\pADFP}{\varavec\vard\varf^\top\varp^top}

\newcommand{\normdist}[1]{\mathcal{N}(\bm{#1}; \musubbf{#1} , \Sigmasubbf{#1})}
\newcommand{\normdistp}[2]{\mathcal{N}(#1,#2)}

\newcommand{\gammadist}[2]{\text{Gamma}(#1,#2)}

\def\vMF{\text{vMF}} 
\newcommand{\matrixnormdistp}[3]{\mathcal{MN}\left(#1,#2,#3\right)}

\newcommand{\pDist}[1]{\text{p}(#1)}
\newcommand{\pDistCon}[2]{\text{p}(#1\mid#2)}
\newcommand{\qDist}[1]{\text{q}(#1)}

\newcommand{\zbf}{\varparameters} %
\newcommand{\xbf}{\bm{X}} 

\newcommand{\reff}[1]{(\ref{#1})}

\newcommand{\refeq}[1]{(\ref{eq:#1})}

\newcommand{\setfiglabel}[1]{
\@ifundefined{r@fig:#1}{%
    \label{fig:#1}%
  }{%
    \label{fig:#1_appendix}%
  }%
}

\newcommand{\updateAsim}[0]{
	\qDist{\vara} \sim \prod\limits_i\normdistp{\musub{\varavec}}{\Sigmasub{\varavec}}
}

\newcommand{\updateAmean}[0]{
\musub{\varavec} = \Sigmasub{\varavec}\sum\limits_\nslabslc \e{\varsigmavec}\e{\vardvec\varf^\top\varp^\top\varxvec^\top}
}

\newcommand{\updateAvar}[0]{
\Sigmasub{\varavec} = \inverse{\big(\Identity{\ncomps}+\sum\limits_\nslabslc\e{\varsigmavec}\e{\vardvec\varf^\top\varp^\top\varp\varf\vardvec}\big)}
}

\newcommand{\updateCsim}[0]{
	\qDist{\varc} \sim \prod\limits_\nslabslc \normdistp{\musub{\varcvec}}{\Sigmasub{\varcvec}}
}

\newcommand{\updateCmean}[0]{
\musub{\varcvec} = \Sigmasubbf{\varcvec} \e{\varsigmavec}\diag{\e{\varf^\top}\e{\varp^\top}\varx^\top\e{\vara}}
}

\newcommand{\updateCvar}[0]{
	\Sigmasub{\varcvec} = \inverse{\big(\diag{\e{\varalpha}}+\e{{\varsigmavec}}\e{\varf^\top\varp^\top\varp\varf}* \e{\vara\tran\vara}\big)}
}

\newcommand{\updateFsim}[0]{
	\qDist{\varf} \sim \prod\limits_\ncompslc \normdistp{\musub{\varfvec}}{\Sigmasub{\varfvec}}
}

\newcommand{\updateFmean}[0]{
\musubbf{\varfvec} = \Sigmasubbf{\varfvec}\big(\sum\limits_\nslabslc\e{\varsigmavec}
\big\{\e{(\varp^\top)_\ncompslc}\varx^\top\e{\vara}\e{\vardvec}-\e{\vardvec\vara\tran\vara\vardvec}\sum\limits_{\ncompslc\prime \setminus \ncompslc} \e{\bm{p}^\top_{\cdot \ncompslc \nslabslc}\bm{p}_{\cdot \ncompslc\prime \nslabslc}}\bm{f}_{\ncompslc\prime\cdot}^\top\big\}\big)
}

\newcommand{\updateFvar}[0]{
\Sigmasubbf{\varfvec} = \inverse{\big(\sum_\nslabslc\e{\varsigmavec}\e{\vardvec\vara\tran\vara\vardvec}\e{\bm{p}^\top_{\cdot \ncompslc \nslabslc}\bm{p}_{\cdot \ncompslc \nslabslc}}+\Identity{\ncomps}\big)}
}

\newcommand{\updatePsim}[0]{
	\qDist{\fullp} \sim \prod\limits_{\nslabslc} c\matrixnormdistp{\Msub{\varp}}{\Identity{J}}{\Sigmasub{\varp}}
}

\newcommand{\updatePmean}[0]{
	\begin{aligned} 
		\Msub{\varp} = 
        \bm{V}_\nslabslc\bm{U}_\nslabslc^\top, \text{where}\  \e{\varsigmavec}
        \e{\varf}\e{\vard}\e{\vara^{\!\top}}\varx = \bm{U}_\nslabslc\bm{S}_\nslabslc\bm{V}_\nslabslc^\top \text{(SVD)}
\end{aligned}
}

\newcommand{\updatePvar}[0]{
	\Sigmasub{\varp} = 
    	\inverse{(\e{\varf\vardvec\vara\tran\vara\vardvec\transpose\varf}+\Identity{})}
}

\newcommand{\updatePsimVMF}[0]{
	\qDist\fullp \sim \prod\limits_k\vMF(\bm{B}_{\varp})
}

\newcommand{\updatePparVMF}[0]{
	\bm{B}_{\varp} = \e{\varsigmavec}\e{\varf}\e{\vard}\e{\vara^{\!\top}}\varx
}

\newcommand{\updatePmeanVMF}[0]{
	\begin{aligned} 
		\e\varp = \bm{V}_\nslabslc\bm{\Psi} \transpose{\bm{U}}_\nslabslc\text{, where}\ \bm{B}_{\varp} = \bm{U}_\nslabslc \bm{S}_\nslabslc \transpose{\bm{V}}_\nslabslc \text{(SVD)}
\end{aligned}
}

\newcommand{\updateNoisesim}[0]{
\qDist{\varsigma} \sim \prod\limits_\nslabslc \gammadist{a_{\varsigmavec}}{b_{\varsigmavec}}
}

\newcommand{\updateNoisealpha}[0]{
a_{\varsigmavec} = \varsigmaa+\frac{I\cdot J}{2}
}

\newcommand{\updateNoisebeta}[0]{
b_{\varsigmavec} = \inverse{\big(\inverse\varsigmab\!+\!\frac{1}{2}\trace{\varx\varx^\top}\!+\!\frac{1}{2}\e{\trace{\vara\vardvec\varf^\top\varp^\top\varp\varf\vardvec\vara^{\!\top}}} -\e{\trace{\vara\vardvec\varf^\top\varp^\top\varx^\top}}\big)}
}

\usepackage{rotating}
\usepackage{array}

\makeatletter 
\if@twocolumn 

\else

\fi 
\makeatother

\usepackage{pdflscape}

\usepackage{amsmath,amsthm,amssymb}

\usepackage{bm}
\usepackage{url}
\usepackage{nomencl}
\makenomenclature

\usepackage{todonotes,varwidth}
\makeglossaries
\newglossaryentry{test term}
{
  name=test term,
  description={test description},
}

\newacronym{pca}{PCA}{principal component analysis}
\newacronym{snr}{SNR}{signal-to-noise ratio}
\newacronym{svd}{SVD}{singular value decomposition}
\newacronym{ard}{ARD}{automatic relevance determination}
\newacronym{elbo}{ELBO}{evidence lower bound}
\newacronym{ccd}{CCD}{core consistency diagnostic}
\newacronym{kldiv}{KL divergence}{Kullbacker-Leibler divergence}
\newacronym{parafac2}{}{PARAFAC2}
\newacronym{vmf}{vMF}{Von Mises-Fisher}
\newacronym{cp}{CP}{CandeComp/PARAFAC}

\begin{document}
%
\title{Probabilistic PARAFAC2}

\author{Philip~J.~H.~J\o rgensen$^\dagger$, 
			S\o ren F. V. Nielsen$^\dagger$, 
            Jesper L. Hinrich$^\dagger$, 
            Mikkel N. Schmidt$^\dagger$, 
            Kristoffer H. Madsen$^{\dagger, \ddagger} $, 
            Morten M\o rup$^\dagger$
\thanks{$\dagger$ Department of Applied
Mathematics and Computer Science, Technical University of Denmark, 2800 Kgs. Lyngby, Denmark, $\ddagger$ Danish Research Centre for Magnetic Resonance, Centre for Functional and Diagnostic Imaging and Research,
Copenhagen University Hospital Hvidovre, 2650 Hvidovre, Denmark}%
\thanks{This work is supported by Innovation Fund Denmark through the Danish Center for Big Data Analytics and Innovation (DABAI).}
}

\IEEEtitleabstractindextext{%
\begin{abstract}
The PARAFAC2 is a multimodal factor analysis model suitable for analyzing multi-way data when one of the modes has incomparable observation units, for example because of differences in signal sampling or batch sizes.
A fully probabilistic treatment of the PARAFAC2 is desirable in order to improve robustness to noise and provide a well founded principle for determining the number of factors, but challenging because the factor loadings are constrained to be orthogonal.
We develop two probabilistic formulations of the PARAFAC2 along with variational procedures for inference: In the one approach, the mean values of the factor loadings are orthogonal leading to closed form variational updates, and in the other, the factor loadings themselves are orthogonal using a matrix Von Mises-Fisher distribution. We contrast our probabilistic formulation to the conventional direct fitting algorithm based on maximum likelihood. 
On simulated data and real fluorescence spectroscopy and gas chromatography-mass spectrometry data, we compare our approach to the conventional PARAFAC2 model estimation and find that the probabilistic formulation 
is more robust to noise and model order misspecification.
The probabilistic PARAFAC2 thus forms a promising framework for modeling multi-way data accounting for uncertainty. 
\end{abstract}

}

\maketitle

%
\IEEEpeerreviewmaketitle

\IEEEraisesectionheading{\section{Introduction}}

 \IEEEPARstart{T}{ensor} decompositions are multi-way generalizations of matrix decompositions such as \gls{pca}: A matrix is a second order array with two modes, rows and columns, while a data cube is a third order array with the third mode referred to as slabs. When multi-way data has inherent multi-linear structure, the advantage of tensor decomposition methods is that they capture this intrinsic information and often provide a unique representation without needing further constraints such as sparsity or statistical independence.

Tensor factorization originated within the field of psychometrics~\cite{Carroll1970,Harshman1970}, and has proved widely useful in other fields such as chemometrics~\cite{bro1997parafac} for example to model the relationship between excitation and emission spectra of samples of different mixed compounds obtained by fluorescence spectroscopy~\cite{Appellof1981-na}. Tensor decomposition is today encountered in practically all fields of research including signal processing, neuroimaging, and information retrieval (see also~\cite{kolda2009tensor,morup2011applications}).

The two most prominent tensor decompositions are i)~the Tucker model \cite{tucker1966some}, where the so-called core array accounts for all multi-linear interactions between the components of each mode, and ii)~the \gls{cp} model \cite{Carroll1970,Harshman1970, hitchcock1927expression}, where interactions are restricted to be between components of identical indices across modes, corresponding to a Tucker model with a diagonal core array. Both models can be considered generalizations of \gls{pca} to higher order arrays, with the Tucker model being more flexible at the expense of reduced interpretability. The CP model has been widely used primarily due to its ease of interpretation and its uniqueness~\cite{morup2011applications}.

In the CP model the components are assumed identical across measurements, varying only in their scaling. In many situations this is too restrictive, for example when signal sampling or batch sizes vary across a mode. In chemometrics, violation of the CP structure can be caused by retention time shifts~\cite{bro1999parafac2}, whereas in neuroimaging violation can be caused by subject and trial variability~\cite{morup2011applications}. To handle variability while preserving the uniqueness of the representation, the PARAFAC2 model was proposed \cite{Harshman1970}. It admits individual loading matrices for each entry in a mode while preserving uniqueness properties of the decomposition by imposing consistency of the Gram matrix (i.e. the loading matrix left multiplied by its transpose)~\cite{harshman1996uniqueness,ten1996some,kiers1999parafac2}. It has since been applied within diverse application domains such as in chemometrics for handling variations in elution profiles due to retention shifts in chromatography~\cite{bro1999parafac2}, for monitoring and fault detection facing unequal batch lengths in chemical processes~\cite{wise2001application}, in neuroimaging to analyze latency changes in frequency resolved evoked EEG potentials~\cite{weis2010multi}, to extract common connectivity profiles in multi-subject fMRI data accounting for individual variability~\cite{madsen2016quantifying}, for cross-language information retrieval~\cite{chew2007cross}, and for music and image tagging~\cite{panagakis2011automatic,pantraki2015automatic}. Recently, efforts have been made to scale the PARAFAC2 model to large-scale data \cite{Tian2018-ah, Perros2017-ti} and a nonnegative version have been developed \cite{Cohen2018-kh}.

Traditionally, tensor decompositions have been based on maximum likelihood inference using alternating least squares estimation in which the components of a mode are estimated while keeping the components of other modes fixed. Initial probabilistic approaches defined probability distributions over the component matrices and the core array, but relied on maximum likelihood estimates for determining a solution. However, the Bayesian approach presented here makes inference with respect to the posterior distributions of the model parameters, and can thus be used to asses uncertainty in the parameters and noise estimates. Recently, the TUCKER and CP models have been formulated in a probabilistic setting, using either Markov Chain Monte Carlo (MCMC) sampling~\cite{porteous2008multi,sheng2012probabilistic,bhattacharya2011sparse} or variational inference~\cite{shan2011probabilistic, Ermis2014, xu2012infinite,zhao2015bayesian}. The \gls{cp} and Tucker models have been extended to model sparsity~\cite{hore2016tensor,beliveau2016spparafac,bhattacharya2011sparse}, non-negativity~\cite{Schmidt2009-xs} and non-linearity~\cite{porteous2008multi,xu2015bayesian} in component loadings. Heteroscedastic noise modeling has been discussed in the context of the \gls{cp} model~\cite{beliveau2016spparafac,zhao2016bayesian} and Tucker model~\cite{hayashi2012exponential}, the latter also providing a generalization of tensor decomposition to exponential family distributions. A \gls{cp} model where a subset of the component matrices are orthogonal matrices was recently explored using the von-Mises-Fisher distribution~\cite{cheng2017probabilistic}, their approach was not fully Bayesian, as they used MAP estimates for the orthogonal matrices, neither did they explore other orthogonal formulation or the PARAFAC2 structure.

Benefits of probabilistic modeling include the ability to account for uncertainty and noise while providing tools for model order selection. Whereas probabilistic modeling can be directly applied to the CP and TUCKER models extending probabilistic PCA~\cite{Bishop1999}, a probabilistic treatment of the PARAFAC2 model faces the following two key challenges, i) the ability to impose orthogonality on variational factors (necessary for imposing the PARAFAC2 structure), and ii) handling the coupling of these orthogonal components. In this paper we address these challenges and derive the probabilistic PARAFAC2 model. We investigate two different formulations of the orthogonality constraints and demonstrate how orthogonality of variational factors in least squares estimation as for conventional PARAFAC2 can be obtained in closed form using the singular value decomposition. We exploit how the probabilistic framework admits model order quantification by the evaluation of model evidence and relevance determination. We contrast our probabilistic formulation to conventional maximum likelihood estimation on synthetic data as well as flourescence spectroscopy and gas chromatography-mass spectrometry data highlighting the utility of the probabilistic formulation facing noise and model order misspecification.
 
\section{Methods}
\label{sec:methods}
The three-way \gls{cp} model can be formulated as a series of coupled matrix decompositions, 
\begin{align}
	\varx &= \vara\vard\transpose\varf+\vare,
    \label{eq:parafac}
\end{align}
where $\varx \in \mathbb{R}^{I \times J}$ is the $k$'th slab of the three-way array $\fullx$ with dimensions ${I\times J\times K}$. Let $\ncomps$ be the number of components in the model, then the matrix $\vara$ with dimensions $I \times \ncomps$ contains loadings for the first mode and $\varf$ with dimensions $J \times \ncomps$ contains loadings for the second mode. The matrices $\vard$, $k=1,\hdots,K$, are diagonal with dimensions $\ncomps \times \ncomps$ and contain loadings for the third mode. These are usually stored as a single matrix $\varc \in \mathbb{R}^{K \times \ncomps}$ where the $k$'th row contains the diagonal of $\vard$. $\vare$ denotes the residuals for the $k$'th slab with dimensions $I \times J$. Notice that the structure of the first and second mode are invariant across the third mode in this model. 

The \acrlong{parafac2} model extends the \gls{cp} structure by letting a mode have individual factors $\varf_k$ for each slab. The extension allows for a varying number of observations in the chosen mode. This model would be as flexible as PCA on the concatenated data $[\bm{X}_1, \bm{X}_2, \ldots \bm{X}_K]$ if not for the additional constraint that each Gram matrix of $\varf_k$ be identical, $\varf_k^\top\varf_k = \Psi$ which is a necessary constraint in order to obtain unique solutions \cite{Harshman1972}. The three-way PARAFAC2 model can thus be written as,
\begin{align}
	\varx &= \vara\vard\varf_k^\top+\vare\quad \text{s.t.}\enspace\varf_k^\top\varf_k = \Psi.
    \label{eq:parafacpsi}
\end{align}
Modeling $\Psi$ explicitly can be difficult, but it is necessary and sufficient~\cite{kiers1999parafac2} to have $\varf_k = \varp\varf$, with $\varp$ a $J \times \ncomps$ columnwise orthonormal matrix, and $\varf$ a $\ncomps \times \ncomps$ matrix, and the model can thus be written as
\begin{align}
	\varx &= \vara\vard\transpose{\varf}\varp^\top + \vare \text{   s.t.   } \varp^\top\varp = \Identity{}. \label{eq:parafac2}
\end{align}
In the following, we describe a direct fitting algorithm \cite{kiers1999parafac2} for parameter estimation in the \acrlong{parafac2} model, before we introduce the probabilistic model formulation.

\subsection{Direct Fitting Algorithm Using Alternating Least Squares}
The parameters in the PARAFAC2 model in \eqref{eq:parafac2} can be estimated using the alternating least squares algorithm \cite{kiers1999parafac2}, minimizing the constrained least squares objective function,
\begin{align}
\min \sum_k \|\bm{X}_k&-\bm{AD}_k\bm{\transpose{F}}\bm{P}_k^\top\|^2 
\label{eq:directfitobjective} \\
 \mathrm{s.t.}\quad \bm{P}_k^\top \bm{P}_k &= \bm{I}. \nonumber
\end{align}
For fixed $\vara$, $\vard$, and $\varf$, the $\varp$ that minimizes the $k$'th term in the objective function is equal to the $\varp$ that maximizes
\begin{align}
	\trace{\bm{F}\bm{D}_k\bm{A}^{\!\top}\bm{X}_k\varp}, \label{eq:parasvd}
\end{align}
and can be computed as \cite{Green1952,kiers1999parafac2}
\begin{align}
\varp = \bm{V}_k\bm{U}_k^\top, \label{eq:maxPk}
\end{align}
where $\bm{V}_k$ and $\bm{U}_k$ comes from the \gls{svd} decomposition 
\begin{equation} \bm{U}_k\boldsymbol{S}_k\bm{V}_k^\top=\bm{F}\bm{D}_k\bm{A}^{\!\top}\bm{X_k}.
\end{equation} 
Upon fitting $\varp$ each slab $\bm{X}_k$ of the tensor can be projected onto $\varp$ thereby leaving the remaining parameters to be fitted as a \gls{cp} model minimizing 
\begin{equation} 
	\sum\limits_k\|\varx\varp-\bm{AD}_k\bm{\transpose{F}}\|^2. \label{eq:para2para}
\end{equation}
A solution to \eqref{eq:para2para} is well explained by Bro in \cite{Bro1997-li}. A well-known issue with maximum likelihood methods is that it can lead to overfitting due to noise and a lack of uncertainty in the model parameters resulting in robustness issues. These problems we hope to solve by advancing the \acrlong{parafac2} model to a fully Bayesian setting.

\subsubsection{Model Selection}
A general problem for latent variable methods is how to choose the model order. A popular heuristic would be how well the model fits the data given asKig på AIS Data
\begin{align}
\text{R2} = 1-\frac{\sum\limits_k\|\varx-\bm{AD}_k\bm{\transpose{F}}\bm{P}_k^\top\|^2.}{\sum\limits_k\|\varx\|^2.} \label{eq:fitpct}
\end{align}
However, this measure will simply increase until the model incorporates enough parameters to completely fit the data, thus eventually leading to overfitting. The model selection criteria would only be based on the expected noise level. 

Another popular heuristic is the \gls{ccd} originally developed for the \gls{cp} model\cite{bro2003new}, but shown useful for the \acrlong{parafac2} model as well\cite{Kamstrup-Nielsen2013}. It is based on the observation that the PARAFAC model can been seen as a constrained Tucker model, where the core array is enforced to be a superdiagonal array of ones. The principle behind \gls{ccd} is to measure how much the PARAFAC model violates this assumption of a superdiagonal core array of ones by re-estimating the core array of the PARAFAC model to fit the Tucker model, denoted $\mathcal{G}$, while keeping the loadings fixed and then calculating the \gls{ccd} according to,
\begin{align}
	\text{\gls{ccd}} = 100 \left( 1 - \frac{|| \mathcal{G} - \mathcal{I} ||^2_{\mathcal{F}}}{||\mathcal{I} ||^2_{\mathcal{F}}} \right), 
\end{align}
in which $\mathcal{I}$ is the superdiagonal core array and $||\cdot||_{\mathcal{F}}$ denotes the Frobenius norm. The \acrlong{parafac2} model can be written as a PARAFAC model for each slab as in \eqref{eq:para2para}, and thus the core array can be estimated in the same way as for the standard PARAFAC model. This approach have been evaluated on simulated as well real data sets by Kamstrup-Nielsen etKig på AIS Data al. \cite{Kamstrup-Nielsen2013-wb} where the conclusion is even though the \gls{ccd} is found to be an useful parameter for determining model order, it is not recommended to be used without considering more diagnostic measures such as residuals and the loadings.

\subsection{Variational Bayesian Inference}
In Bayesian modeling, the posterior distribution of the parameters $\bm{\theta}$ is computed by conditioning on the observed data $\xbf$ using Bayes' rule,
\begin{math}
p(\bm{\theta}|\xbf) = \frac{p(\xbf|\bm{\theta})p(\bm{\theta})}{p(\xbf)}.
\end{math}
It is given by the product of the likelihood $p(\xbf|\bm{\theta})$ and the prior probability of the parameters $p(\bm{\theta})$, divided by the probability of the observed data $p(\xbf)$ under the model, also known as the marginal likelihood. Evaluating the marginal likelihood is in general intractable, and instead a variational approximation can be found by fitting a distribution $q(\bm{\theta})$ to the posterior~\cite{attias1999variational} minimizing the Kullback-Leibler (KL) divergence, given by

\begin{align}
	q^\star(\bm{\zbf}) = 
    \text{arg min KL} \big[q(\zbf) \big\| p(\zbf|\xbf)\big].
\end{align}

Minimizing the KL divergence is solved by maximizing a related quantity, the \gls{elbo}.
\begin{align}
	\text{ELBO}(q(\bm{\theta})) = \e{\log p(\zbf,\xbf)}-\e{\log q(\zbf)}. \label{eq:elbo}
\end{align}
A common choice is a variational distribution that factorizes over the parameters, known as a mean-field approximation, $q(\bm{\theta})=\prod_j q_j(\bm{\theta}_j)$. The optimal variational distribution can then be found by iterative updates of the form
\begin{align}
	q_j(\varfactorj) \propto& 
	\exp \big( \ej{j}{\log p(\varfactorj , \zbf_{-j},\xbf)}\big),
	\label{eq:varfacdef}
\end{align}
where $\ej{j}{\cdot}$ denotes the expectation over the variational distribution except $q_j$. For a comprehensive overview of variational inference see for example \cite{Bishop2006,Blei2016}.
\subsection{Probabilistic PARAFAC2}
We propose a probabilistic PARAFAC2 using the formulation in eq.~\eqref{eq:parafac2}. The constraint $\varp^\top\varp=\bm{I}_M$ has the probabilistic interpretation i) $\e{\varp^\top\varp}=\bm{I}_\ncomps$, but one could also consider to model ii) $\e{\varp}^\top\e{\varp}=\bm{I}_\ncomps$. The main motivation for the latter approach being the interpretation of the orthogonal factor is identical to that of the maximum likelihood estimation, however the resulting components are no longer themselves restricted to the Stiefel manifold. As such, the model becomes more flexible as only the mean parameters of the variational approximation are constrained to be orthogonal, but not the expectation of their inner product being orthogonal as required for every realization of the underlying distribution to conform to the PARAFAC2 model. We included the latter model formulation as it provides simple closed form updates similar to the original PARAFAC2 direct fitting procedure as shown below. The update is derived by constraining the mean of a matrix normal ($\mathcal{MN}$) distribution within the variational approximation to the set of orthogonal matrices (the Stiefel manifold), whereas the former formulation based on \cite{Smidl2007} uses a matrix von Mises-Fisher ($\vMF$) distribution which has support only on the Stiefel manifold. We thus have the generative models i) and ii),

\begin{align*}
	&& \varavec &\sim \normdistp{\bm{0}}{\Identity{\ncomps}}\\
    && \varfvec &\sim \normdistp{\bm{0}}{\Identity{\ncomps}}\\
	&&\varcvec &\sim \normdistp{\bm{0}}{\diag{\varalpha^{-1}}}\\
    \text{i)}&&\varp &\sim \vMF\left(\bm{0}\right)\\
    \text{ii)}&&\varp &\sim \matrixnormdistp{\bm{0}}{\Identity{J}}{\Identity{\ncomps}}\\
    &&\varsigmavec &\sim\!\text{Gamma}(a_{\varsigmavec},b_{\varsigmavec})\\
    &&\varx &\sim \normdistp{\vara\vardvec\varf^\top\varp^\top}{\varsigmavec^{-1}\Identity{J}}
\end{align*}
where $\varavec$ denotes the $i$th row of the matrix $\vara$ etc. In the above formulation, $\boldsymbol{\alpha}$ defines the length scale of each component and can thus be used for automatic relevance determination by turning of excess components by concentrating their distributions at zero when $\alpha_m$ is large\cite{Bishop2006}. We further allow the noise to vary across slabs thereby accounting for potential different levels of the noise (i.e., assuming heteroschedastic noise) across slabs.

\subsection{Variational Update Rules}
The inference is based on the following factorized distribution,
\begin{equation}
	q(\vara) q(\varc) \prod_\ncompslc q(\varfvec)
    \prod_k q(\varp)q(\varsigmavec).
\end{equation}
leading to the following ELBO,
\begin{align}
\begin{split}
	\text{ELBO}(\qDist{ \varparameters })=&\elog{  p(\fullx,\varparameters)}-\elog{  \qDist{ \varparameters } } \\
=&\elog{\pDistCon{\fullx}{\vara,\varc,\varf,\fullp,\varsigma}}+\elog{\pDist{\vara}}\\
	&+\elog{\pDistCon{\varc}{\varalpha}}+\elog{\pDist{\varalpha}}\\
	&+\elog{\pDist{\varf}}+\elog{\pDist{\fullp}}+\elog{\pDist{\varsigma}}\\
	&+\entropy{\qDist{\vara}}+\entropy{\qDist{\varc}}+\entropy{\qDist{\varf}}+\entropy{\qDist{\fullp}}\\
	&+\entropy{\qDist{\varsigma}}+\entropy{\qDist{\varalpha}} \label{eq:PARAFAC2_elbo}
    \end{split}
\end{align}
Using eq.~\eqref{eq:varfacdef}, the resulting variational distributions and update rules are given in Table~\ref{tab:updaterules}. The update for the factor matrix $F$ is non-trivial, and to obtain a closed-form solution we employ a component-wise updating scheme inspired by the non-negative matrix factorization litterature \cite{bro1998multi,gillis2008nonnegative,nielsen2014non}. 
For each latent parameter we use eq.~\eqref{eq:varfacdef} and moment matching to determine the optimal variational distributions.\\

\begin{table*}[t]
\caption{Overview of all the variational factors and their updates.}
\label{tab:updaterules}
\begin{center}
{
	\begin{tabularx}{\textwidth}{>{\centering\arraybackslash} p{\dimexpr0.3\textwidth-2\tabcolsep} m{\dimexpr0.7\textwidth-2\tabcolsep}} 
    \toprule
	Variational factor & Update \\ 
    \midrule
	\multirow{2}{*}{$\updateAsim$} & $\updateAvar$\\
    & $\updateAmean$\\
    \rowcolor[gray]{0.9} 
    \multirow{2}{*}{$\updateCsim$} & $\updateCvar$\\
    \rowcolor[gray]{0.9} 
    	& $\updateCmean$\\
	\multirow{2}{*}{$\updateFsim$} & $\updateFvar$ \\
    	& $\updateFmean$\\
    \rowcolor[gray]{0.9}
	\multirow{2}{*}{$\updatePsimVMF$} & $\updatePparVMF$\\
    \rowcolor[gray]{0.9}
    	& $\updatePmeanVMF$
	 ($\boldsymbol{\Psi}$ given by \cite{Smidl2007}, Appendix A.2)\\ 	
    \multirow{2}{*}{$\updatePsim$}  & $\updatePvar$\hfill\\
	   & $\updatePmean$\\
	\rowcolor[gray]{0.9}  
 	\multirow{2}{*}{$\updateNoisesim$} & $\updateNoisealpha$ \\
	\rowcolor[gray]{0.9}  
		& $\updateNoisebeta$ \\
     $\underset{\varalphavec}{\text{arg max}} ~\text{ELBO}(\varalphavec)$
     	& $\varalphavec = K\inverse{\big(\sum\limits_k\e{\varcele^2}\big)}$ \\ 
	\bottomrule
\end{tabularx}
}
\end{center}
\end{table*}

\subsubsection{Von Mises-Fisher Loading}
In the von Mises-Fisher model for the loading $\varp$, the variational distribution is given by
\begin{equation}
\text{vMF}(\varp|\bm{B}_{\varp}) = \kappa(J,\bm{B}_{\varp}^\top \bm{B}_{\varp})^{-1}\text{exp}\big(\text{tr}[\bm{B}_{\varp}^\top\varp]\big) ,
\end{equation}
which is defined on the Stiefel manifold, ${\varp^\top\varp=\bm{I}}$. The normalization constant is given by ${\kappa={}_0F_1\left(\frac{1}{2}J,\frac{1}{4}\bm{B}_{\varp}^\top\bm{B}_{\varp}\right)}\text{v}_{J,M}$ where $\text{v}_{J,M}$ is the volume of the $J$-dimensional Stiefel manifold described by $M$ components~\cite{khatri1977mises}.

The hypergeometric function with matrix argument can be calculated more efficiently using the SVD of $\bm{B}_{\varp} = \bm{U}_k \bm{S}_k \bm{V}_k^\top$, since  ${}_0F_1\left(\frac{1}{2}J,\frac{1}{4}\bm{B}_{\varp}^\top\bm{B}_{\varp}\right)= {}_0F_1\left(\frac{1}{2}J,\frac{1}{4}\bm{S}_k^2\right)$~\cite{khatri1977mises}. 

Computing expectations over the matrix vMF distribution requires evaluating the hypergeometric function and can be done as described by \cite{Smidl2007}.$^\dagger$\footnote{$\dagger$\
Source code for approximating the hypergeometric function is available online \url{http://staff.utia.cz/smidl/files/mat/OVPCA.zip} (24 Feb 2017). This code was used with default settings and without modifications in the experiments.} Note, that it follows from the vMF matrix distribution that $\e{\varp^\top\varp} = \Identity{}$, but in general $\e{\varp^\top}\e{\varp} \neq \Identity{}$. HoweParafac2ver, if an orthogonal summary representation is desired one can inspect the mode of the vMF given by $U_kV_k^\top$.

\subsubsection{Constrained Matrix Normal Loading}
In the constrained matrix normal model for the loadings $\varp$, instead of using the free form variational approach, we maximize \eqref{eq:PARAFAC2_elbo} as a function of the mean parameter $\Msub{\varp}$ subject to the orthogonality constraint
\begin{math}
\Msub{\varp}\Msub{\varp}^\top = \Identity{\ncomps}.
\end{math}

The constraint consequently causes \eqref{eq:PARAFAC2_elbo} to be constant except for the linear term of the expected log of the probability density function of the data. The reason for this is all other terms do not depend on $\Msub{\varp}$ or only on the matrix product $\Msub{\varp}\Msub{\varp}^\top$, which is equivalent to the identity matrix, resulting in the optimization problem 
\begin{align*}
\underset{\Msub{\varp}}{\text{arg max}}~\text{ELBO}(\Msub{\varp})~\text{s. t.}~\Msub{\varp}\transpose{\Msub{\varp}}= \Identity{},
\end{align*}
where
\begin{align*}
\text{ELBO}(\Msub{\varp}) = ~\sum_k\e{\varsigmavec}\trace{\e{\varf}\e\vardvec\e{\transpose{\vara}}\varx\Msub{\varp}}+\cn{}
\end{align*}

This is equal to eq.~\refeq{parasvd} except for a scalar leading to the same solution as for the maximum likelihood estimation method as given in eq.~\refeq{maxPk}. More details on identifying the expression above are given in the supplemental material. The variance parameter $\Sigmasub{\varp}$ in the variational distribution is optimized using eq.~\refeq{varfacdef}.\\

\subsubsection{The {\bf\emph F} Matrix}
The updates for $\varfvec$ are non-trivial due to an intercomponent dependency. The quadratic term in \eqref{eq:varfacdef} for $\varf$ is

\begin{align*}
&\ej{\varf}{\varavec\vardvec\transpose\varf\transpose{\varp}\varp\varf\vardvec\transpose\varavec} \\
=& \ej{\varf}{\trace{\varf\vardvec\transpose\varavec\varavec\vardvec\transpose\varf\transpose{\varp}\varp}}\\
=& \trace{\varf\ej{\varf}{\vardvec\transpose\varavec\varavec\vardvec}\transpose\varf\ej{\varf}{\transpose{\varp}\varp}}\\
=& \sum_{\ncompslc\ncompslc\prime}(\varf\e{\vardvec\transpose\varavec\varavec\vardvec}\transpose\varf)_{\ncompslc\ncompslc\prime}(\e{\transpose{\varp}\varp})_{\ncompslc\ncompslc\prime}\\
=& \sum_{\ncompslc\ncompslc\prime}\varfvec\e{\vardvec\transpose\varavec\varavec\vardvec}\mathbf{f}_{\ncompslc\prime\cdot}^\text{T}\e{\mathbf{p}^\text{T}_{\cdot \ncompslc k}\mathbf{p}_{\cdot \ncompslc\prime k}}\\
=& \sum_\ncompslc \varfvec\e{\vardvec\transpose\varavec\varavec\vardvec}\e{\mathbf{p}^\text{T}_{\cdot \ncompslc k}\mathbf{p}_{\cdot \ncompslc k}}\mathbf{f}_{\ncompslc\cdot}^\text{T}\\
 &+ 2\sum_\ncompslc\sum_{\ncompslc\prime \setminus \ncompslc} \varfvec\e{\vardvec\transpose\varavec\varavec\vardvec}\e{\mathbf{p}^\text{T}_{\cdot \ncompslc k}\mathbf{p}_{\cdot \ncompslc\prime k}}\mathbf{f}_{\ncompslc\prime\cdot}^\text{T}\\
\end{align*}
where we see that the quadratic term separates into a quadratic and linear part revealing the linear intercomponent dependency. 

\subsubsection{Non-trivial expectations}
Below we detail some non-trivial expectations and the necessary steps to compute them. 

The first group of expectations deals with having the diagonal matrix $\vardvec$ at the start and end of the matrix product sequence for which we want to compute the expectation. The first case is the following expectation
\begin{align*}
\e{\vardvec\transpose\varavec\varavec\vardvec}
\end{align*}
Since having the same diagonal matrix both as the left and the right factor around an inner matrix in a matrix product is equivalent to the Hadamard product of the outer-product of the diagonal with itself and the inner matrix, we can separate the expectation into two parts
\begin{align*}
\e{\vardvec\transpose\varavec\varavec\vardvec} =& \e{\transpose\varcvec\varcvec}\hadamard\e{\transpose\varavec\varavec}.
\end{align*}
where $\varcvec$ is the vector containing the diagonal elements of $\vardvec$. The same rule applies for the following expectation 
\begin{align*}
\e{\vardvec\transpose{\varf}\transpose{\varp}\varp\varf\vardvec}=& \e{\transpose{\varcvec}\varcvec}\hadamard\e{\transpose\varf\transpose{\varp}\varp\varf}. \\
\end{align*}
where the second expectation becomes trivial when using the $\vMF$ prior (ii) as the matrix product $\transpose{\varp}\varp$ is the identity matrix. However, when using the matrix normal distribution (i) we get
\begin{align*}
\e{\transpose{\varp}{\varp}} =& \trace{\Sigmasub{\varp}} + \Identity{\ncomps}. \\
\end{align*}
which lead to the element with index $ij$ of the expectation to be equal to
\begin{align*}
\e{\transpose\varf\transpose{\varp}\varp\varf}_{ij}&=\e{\sum_m (\transpose{\varf})_{im} (\transpose{\varp}\varp\varf)_{mj}} \\
&=\e{\sum_m \transpose{\varf}_{mi} \sum_{m^\prime} (\transpose{\varp}\varp)_{mm^\prime}\varf_{m^\prime j}}\\
&=\sum_m\sum_{m^\prime}  \e{\transpose{\varf}_{mi}\varf_{m^\prime j}}\e{(\transpose{\varp}\varp)_{mm^\prime}}\\
\end{align*}
where since the $m$'th and $m^\prime$ components are independent, we have
\begin{align*}
&\e{\transpose{\varf}_{mi}\varf_{m^\prime j}} =\\ 
&\begin{cases}
\e{\transpose{\varf}_{mi}}\e{\varf_{m^\prime j}} + (\Sigmasubbf{\varfvec})_{ij} & \text{for}~m=m^\prime \\
\e{\transpose{\varf}_{mi}}\e{\varf_{m^\prime j}} & \text{for}~m \neq m^\prime \\
\end{cases}
 \\
\end{align*}
which is the final step to compute this expectation.

These are the most involved expectations required to compute the update rules, where the remaining are either more simple or depends upon the expectations derived here.

\subsection{Noise Modeling}
The probabilistic formulation of PARAFAC2 requires the specification and estimation of the noise precision $\bm \tau$. We presently consider two specifications, i.e. homoscedastic noise in which the noise of each slab $\bm{X}_k$ is identical, i.e. $\tau_1=\ldots=\tau_K$ as assumed in the direct fitting algorithm, and heteroscedatic noise where the model includes a separate precision for each of the $K$ slabs.
\subsection{Model Selection}
A benefit of a fully probabilistic formulation of the PARAFAC2 model is that it provides model order quantification using tools from Bayesian inference. We presently exploit automatic relevance determination (ARD) by learning the length scale $\bm{\alpha}$, see also \cite{bishop2006pattern}. In practice we use the MAP estimates for the ARD because we are more interested in the pruning ability than the uncertainty estimates on $\bm{\alpha}$. If desired, a VB estimate is easily found by letting $\alpha_m$ follow a Gamma distribution, c.f. \cite{Bishop1999}. Finally, the estimated ELBO on the data can also be used to compare different model orders. 
\begin{figure*}[!ht]
    \centering
        \subfloat[simulated data with homoscedastic noise.]{\includegraphics[width=0.48\linewidth,trim={0 0 0 0},clip]{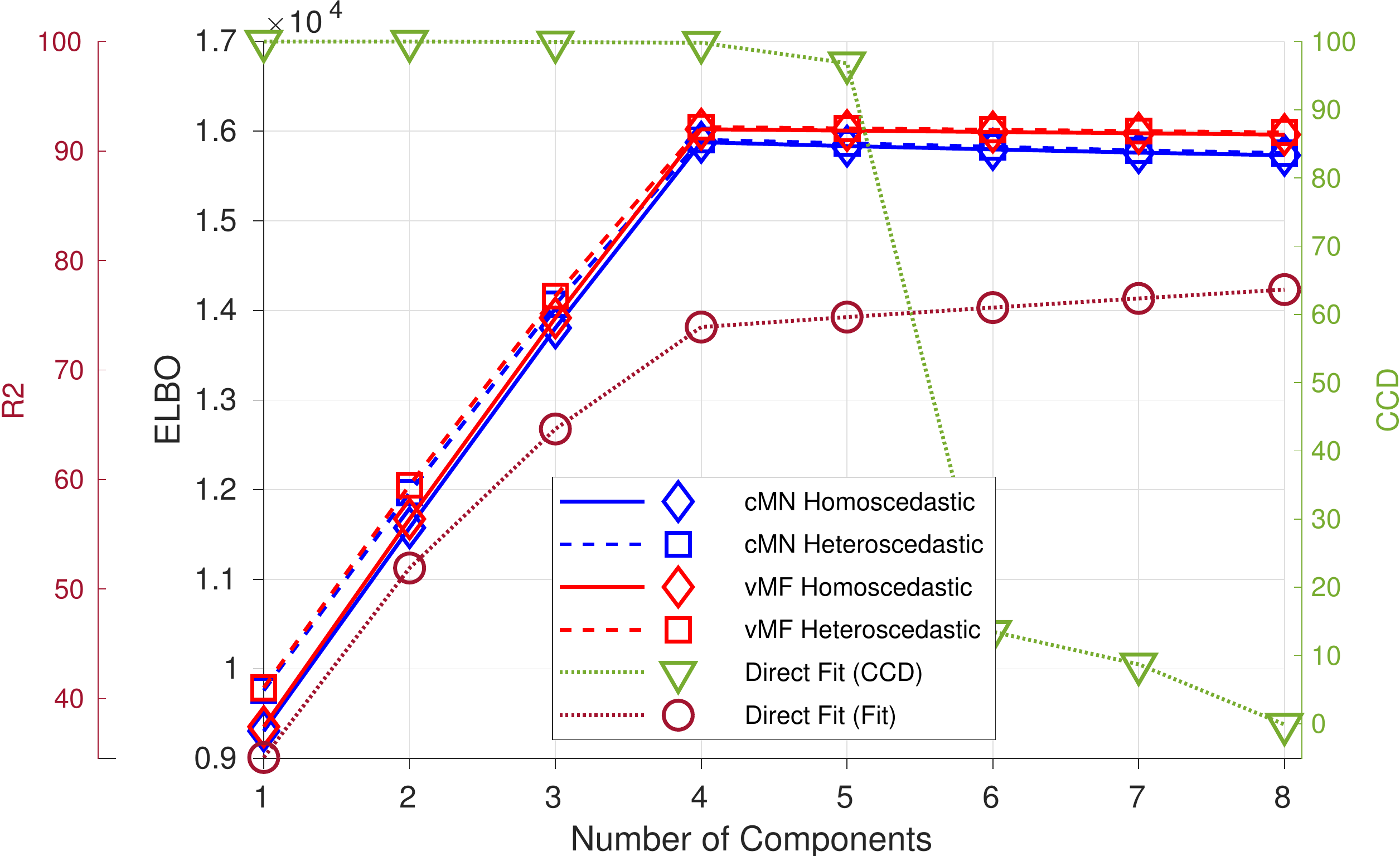}\label{fig:synth_model_order_homo}}~
        \subfloat[simulated data with heteroscedastic noise.]{\includegraphics[width=0.48\linewidth,trim={0 0 0 0},clip]{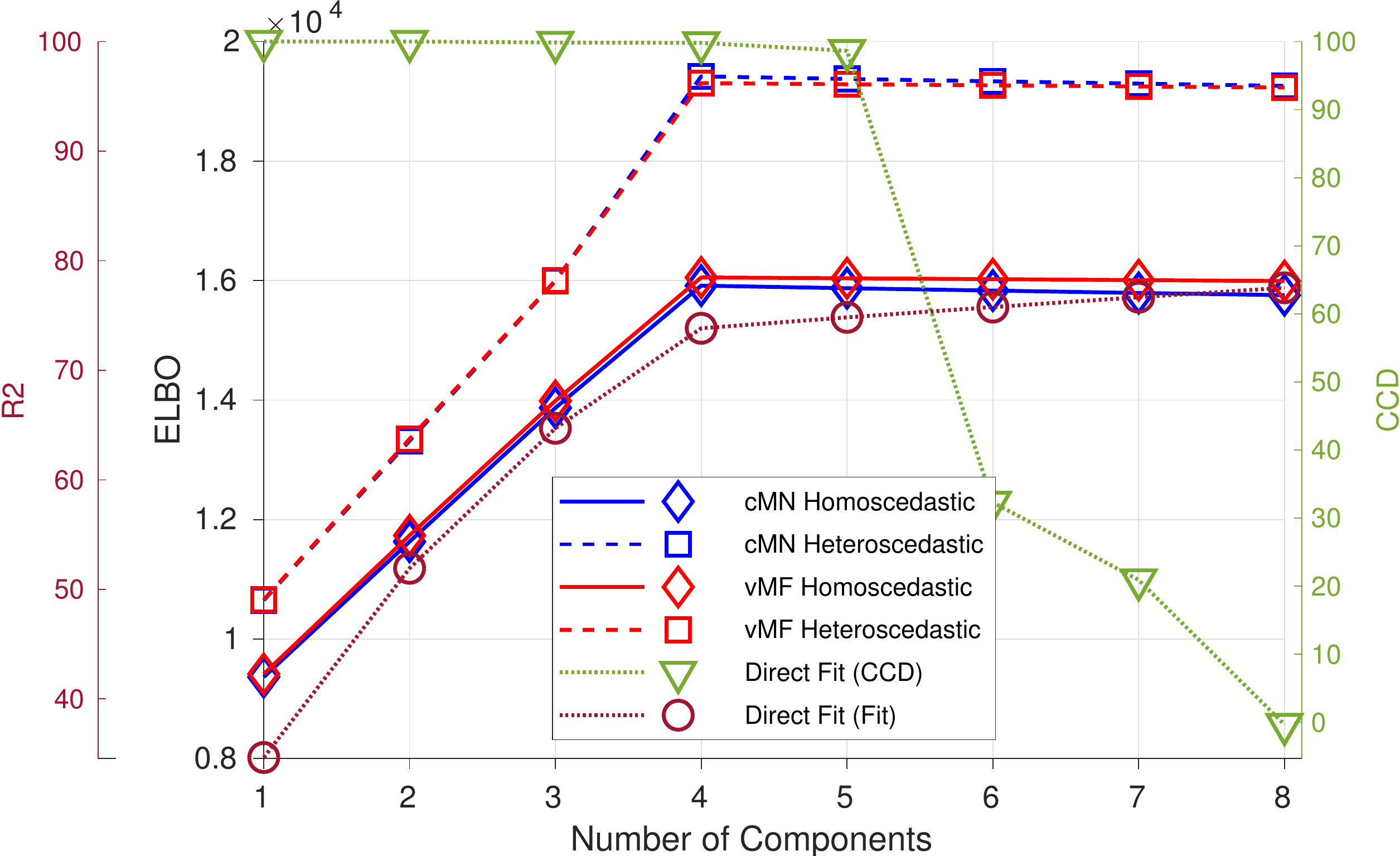}\label{fig:synth_model_order_hetero}}
        \caption{Comparison of model order selection methods in the synthetic data generated with four components and a SNR of 4. We report the different model selection criteria R2, CCD, and ELBO.}
    \label{fig:synth_model_order}
\end{figure*}

\section{Results and Discussion}
In this section we evaluate the proposed model on both synthetic data and two real-world datasets; amino acid fluorescence (AAF) and gas chromatography - mass spectrometry (GC-MS) data. In all cases we initialize the model parameters as the PARAFAC2 solution computed using the direct fitting algorithm\footnote{As implemented by Bro\cite{kiers1999parafac2} at \url{http://www.models.life.ku.dk/go?filename=parafac2.m} (13 Oct 2017)} and perform the model estimation five times to minimize the risk of getting stuck in a local maximum. The final model parameters are chosen as the parameters with the highest ELBO among the five estimations. Each model estimation is limited to $10^4$ iterations or when the change in the ELBO between updates goes below $10^{-9}$. The direct fitting algorithm are also limited to $10^4$ iterations or when the change in R2 \eqref{eq:fitpct} between updates goes below $10^{-12}$. Empirically we experienced better performance when keeping the noise variance fixed for some number of iterations while estimating the length scale $\varalpha$. We choose this delay to last for the first 50 iterations. The hyper-parameters of the precision was set to $(a_{\varsigmavec},b_{\varsigmavec})=(1,10^{32})$ in order to be uninformative for the variational distribution.

\subsection{Simulated Data}
To investigate the performance of the proposed model, we generate simulated datasets in a similar manner as in \cite{Kiers1999}. We generated the data tensor $\fullx$ by sampling $\vara$ from a standard multivariate normal distribution. $\varf$ was taken from a Cholesky factorization of a matrix with $1$'s in its diagonal and $0.4$ in all the off-diagonal elements. This essentially keeps the $\ncomps$ components from being too similar. Each element of $\varc$ was sampled from a uniform distribution on the interval $0$ to $30$. $\varp$ was constructed by the standard orthonormalization function in MATLAB of a set of vectors sampled from a multivariate normal distribution. The simulated datasets were generated with either homoscedastic or heteroscedastic additive noise, at different signal-to-noise ratios (SNR) in the interval $[-20,10]$ with a step of 2. Each configuration was generated 10 times, resulting in 320 data sets. Each data set was given the dimensions $50 \times 50 \times 10$ with $4$ components.

\begin{figure*}[!ht]
    \centering
        \subfloat[Homoscedastic noise with correct model order.]{\includegraphics[width=0.5\linewidth]{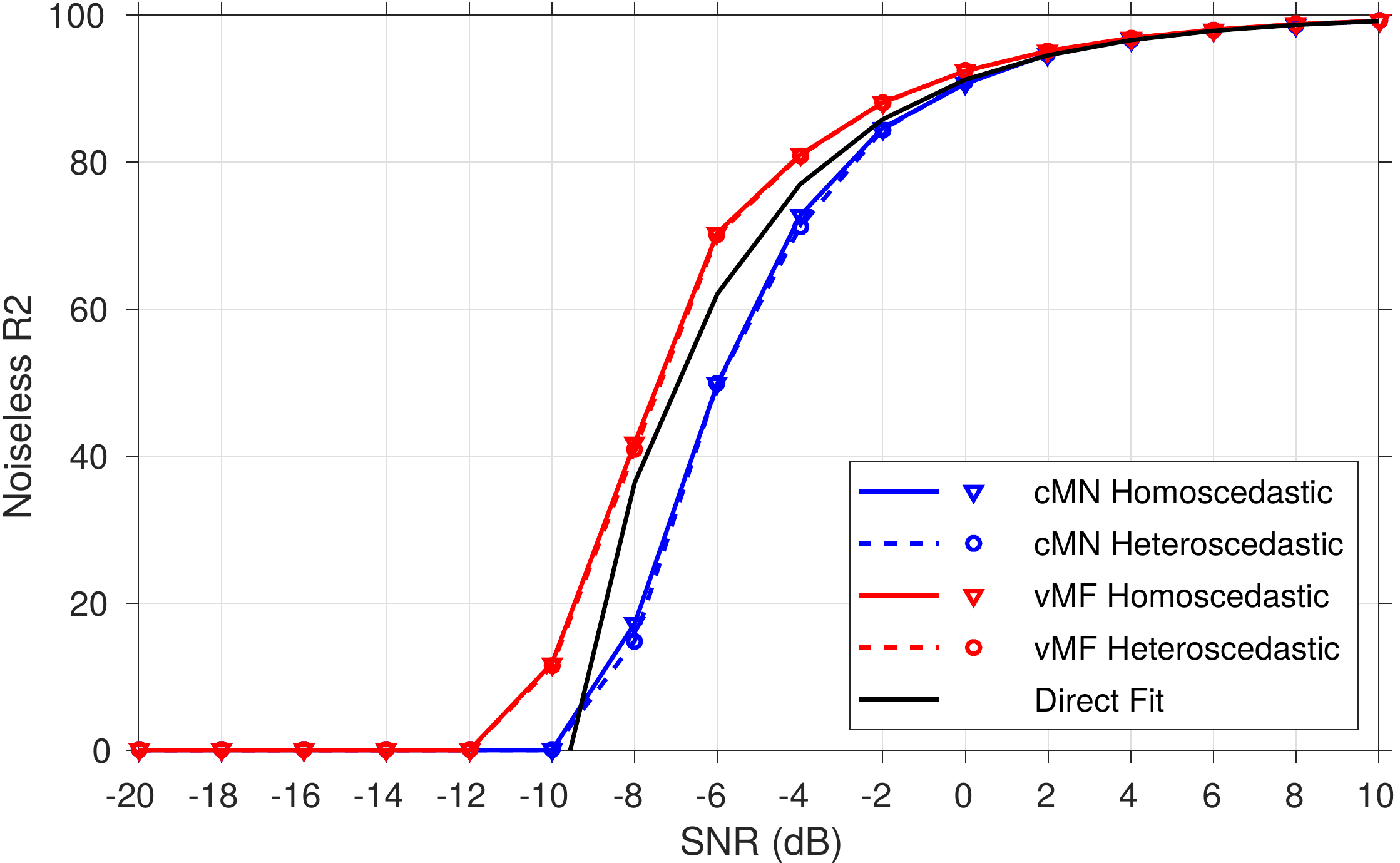}\label{fig:synth_snr_homo}}
        \subfloat[Heteroscedastic noise with correct model order.]{\includegraphics[width=0.5\linewidth]{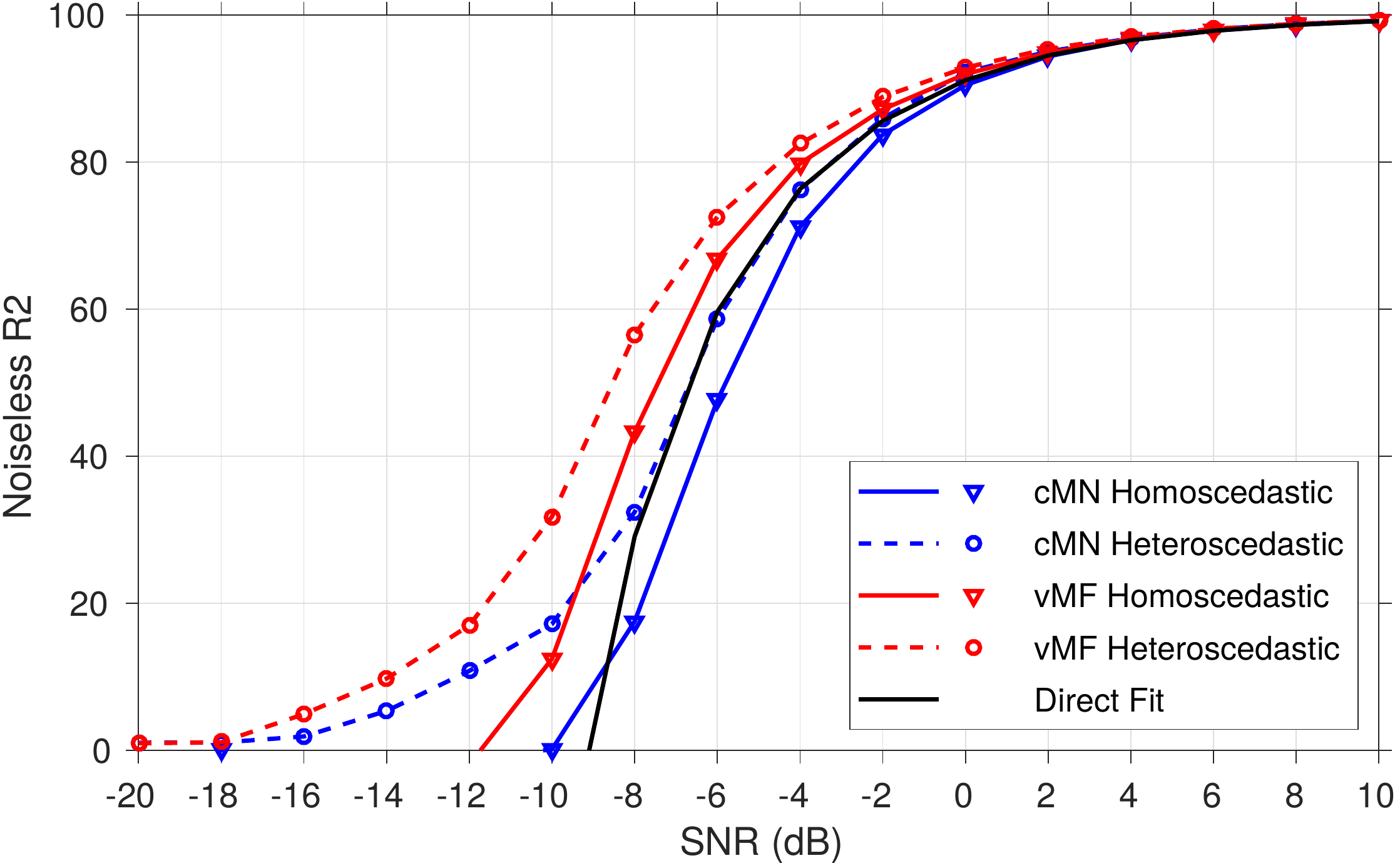}\label{fig:synth_snr_hetero}}
        \qquad
         \subfloat[Homoscedastic noise with incorrect model order.]{\includegraphics[width=0.5\linewidth]{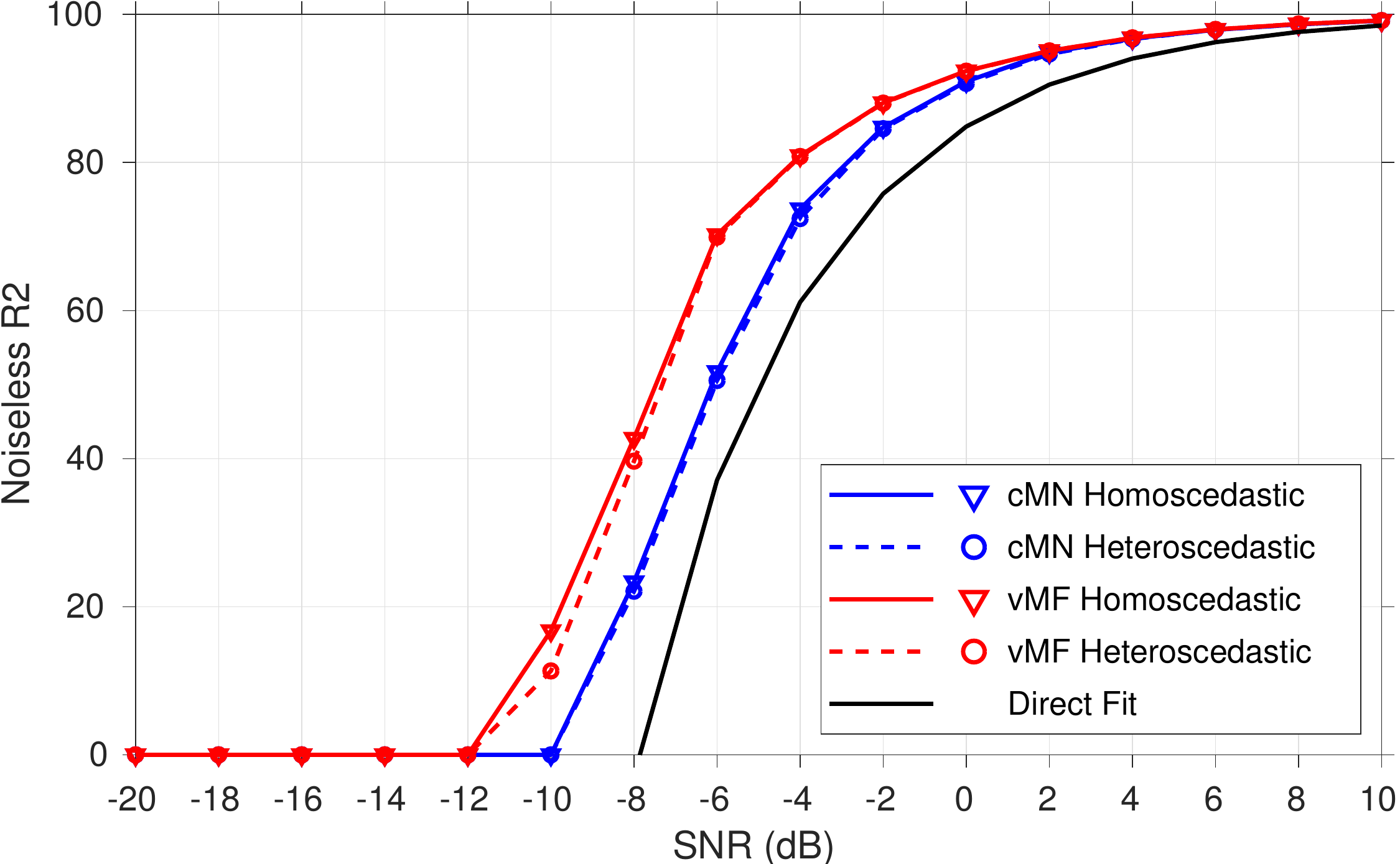}\label{fig:synth_snr_homo_toomany}}
        \subfloat[Heteroscedastic noise with incorrect model order.]{\includegraphics[width=0.5\linewidth]{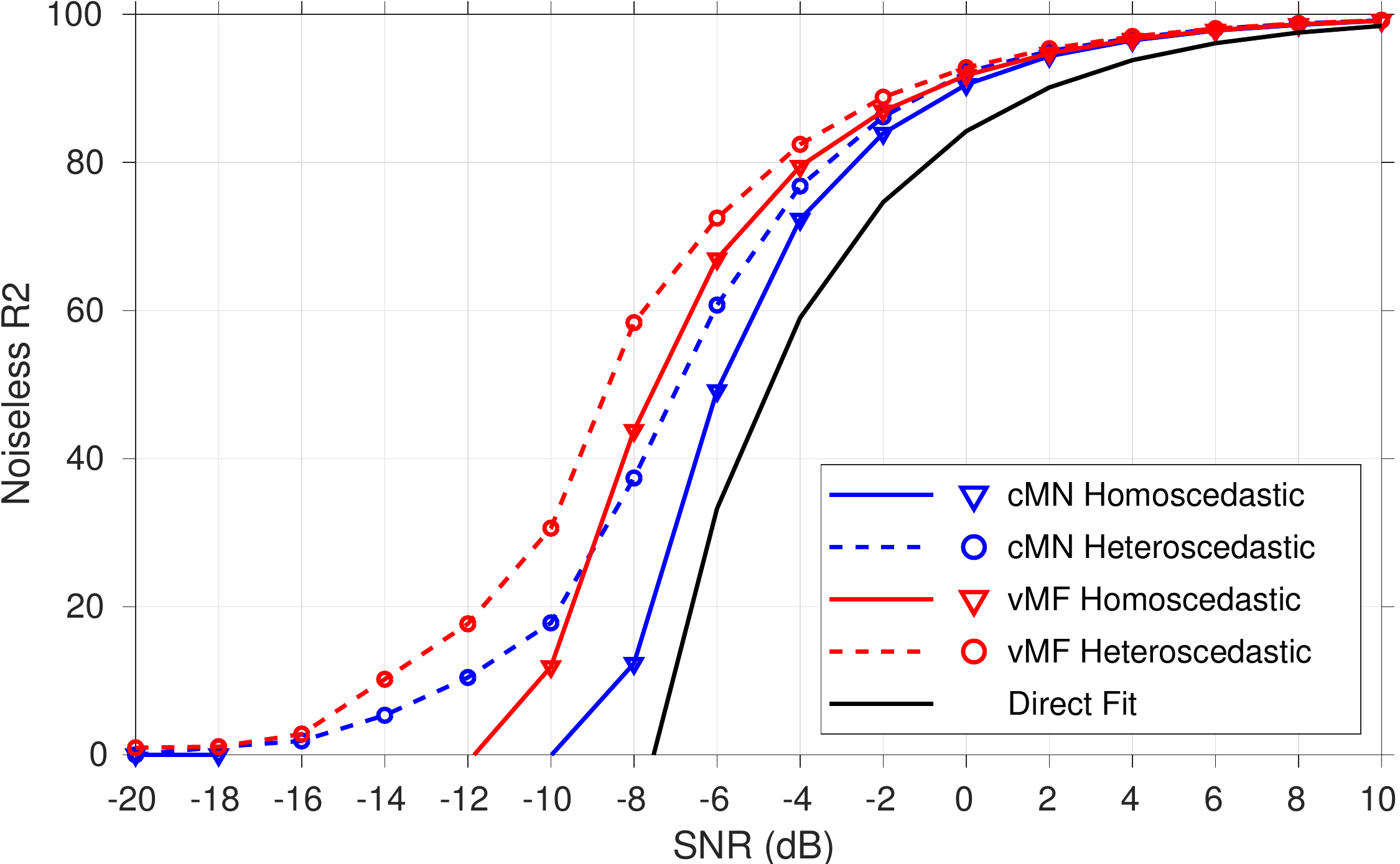}\label{fig:synth_snr_hetero_toomany}}
        \caption{Results from experiment on the generated synthetic data with homoscedastic and heteroscedastic noise respectively. Comparing R2 when computed for the data without the generated noise added for the proposed VB-PARAFAC2 models and the PARAFAC2 fitted on the data with noise added with the correct number of components ($R=4$) shown in the first row, and with too many components ($R=6$) shown in the second row.} 
        \label{fig:synth_snr}
\end{figure*}

The probabilistic PARAFAC2 model was fitted to the datasets, either both the von-Mises-Fisher matrix distribution (vMF) or the constrained mean matrix normal distribution (cMN) for enforcing orthogonality on $\varp$. All results on synthetic data can be seen in Figure~\ref{fig:synth_model_order} and \ref{fig:synth_snr}. We report the R2 on the noise-less data, i.e. using the formula from \eqref{eq:fitpct} with the modification that the noise $\vare$ has been subtracted from $\varx$ for each slab. Thereby, we measure the different models' ability to capture the true underlying structure in the data.

The results for varying SNR using the true number of components in each model are shown in Figure~\ref{fig:synth_snr_homo} for data with homoscedastic noise and in Figure~\ref{fig:synth_snr_hetero} for data with heteroscedastic noise. On the homoscedastic data we see a small advantage of using the two vMF-models over the direct fitting algorithm when we decrease the SNR of the data. The cMN performs slightly worse compared to the direct fitting algorithm. When we move to the the heteroscedastic data, we see a stronger separation of the four different Bayesian methods. Naturally, the models with heteroscedastic noise outperform the ones with homoscedastic noise. It is also evident that the penalty of modelling the noise as heteroscedastic even though it is homoscedastic is small.

Furthermore, if the number of components is misspecified (cf. Figure~\ref{fig:synth_snr_homo_toomany} and Figure~\ref{fig:synth_snr_hetero_toomany}),  we see a larger difference between the performance of the Bayesian models accounting for the heteroscedastic noise and the direct fitting algorithm. The vMF-model again gives better performance compared to the cMN parameterization and we see a larger positive effect of using the Bayesian models over the direct fitting algorithm. This is mainly explained by the reduced tendency to overfit when accounting for the uncertainty and the automatic relevance determination's ability to prune irrelevant components as the Bayesian modeling promotes the simplest possible representation. 

The VB-framework we use for inference gives us a natural way to do model selection through the ELBO. To compare this to the existing model order selection heuristics we plot the different selection criteria as a function of the number of components used in the model in Figure~\ref{fig:synth_model_order_homo} and \ref{fig:synth_model_order_hetero}. The plot is based on synthetic data with four components and an SNR of 4. Overall the ELBO suggests the same number of components as the other two criteria, R2 and \gls{ccd}. Furthermore, when the data has heteroscedastic noise the two Bayesian models that incorporate this have a substantially higher ELBO compared to the homoscedastic models. 
%

\subsection{Real Data}
As our synthetic results pointed to both formulations of the orthogonality constraint to be reasonable, we investigate their performance on two real world data sets. The first dataset is an amino acid fluorescence data\footnote{Available at \url{http://www.models.life.ku.dk/Amino_Acid_fluo}} described in \cite{bro1998multi,kiers1998three} in which the core-consistency diagnostic based on the PARAFAC2 model previously has successfully identified the three underlying constituents; tyrosine, tryptophan and phenylalanine \cite{Kamstrup-Nielsen2013}. The dataset contains five samples with 201 emission and 61 excitation intervals. 

\begin{figure*}[!ht]
    \centering
        \subfloat[AAF data.]{\includegraphics[width=0.48\linewidth]{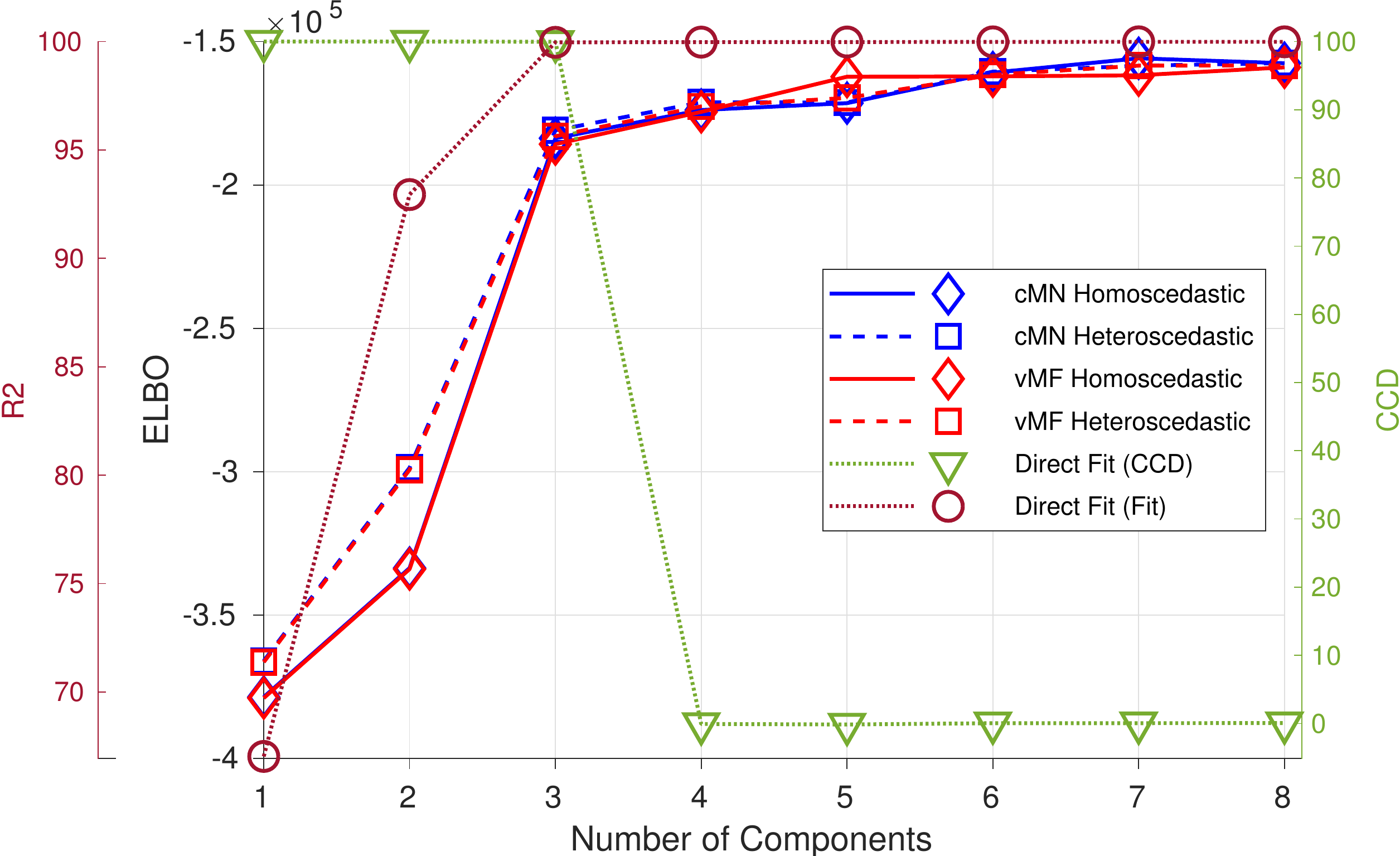}
\label{fig:real_aaf_model_selection}}~
        \subfloat[GC-MS data.]{\includegraphics[width=0.48\linewidth]{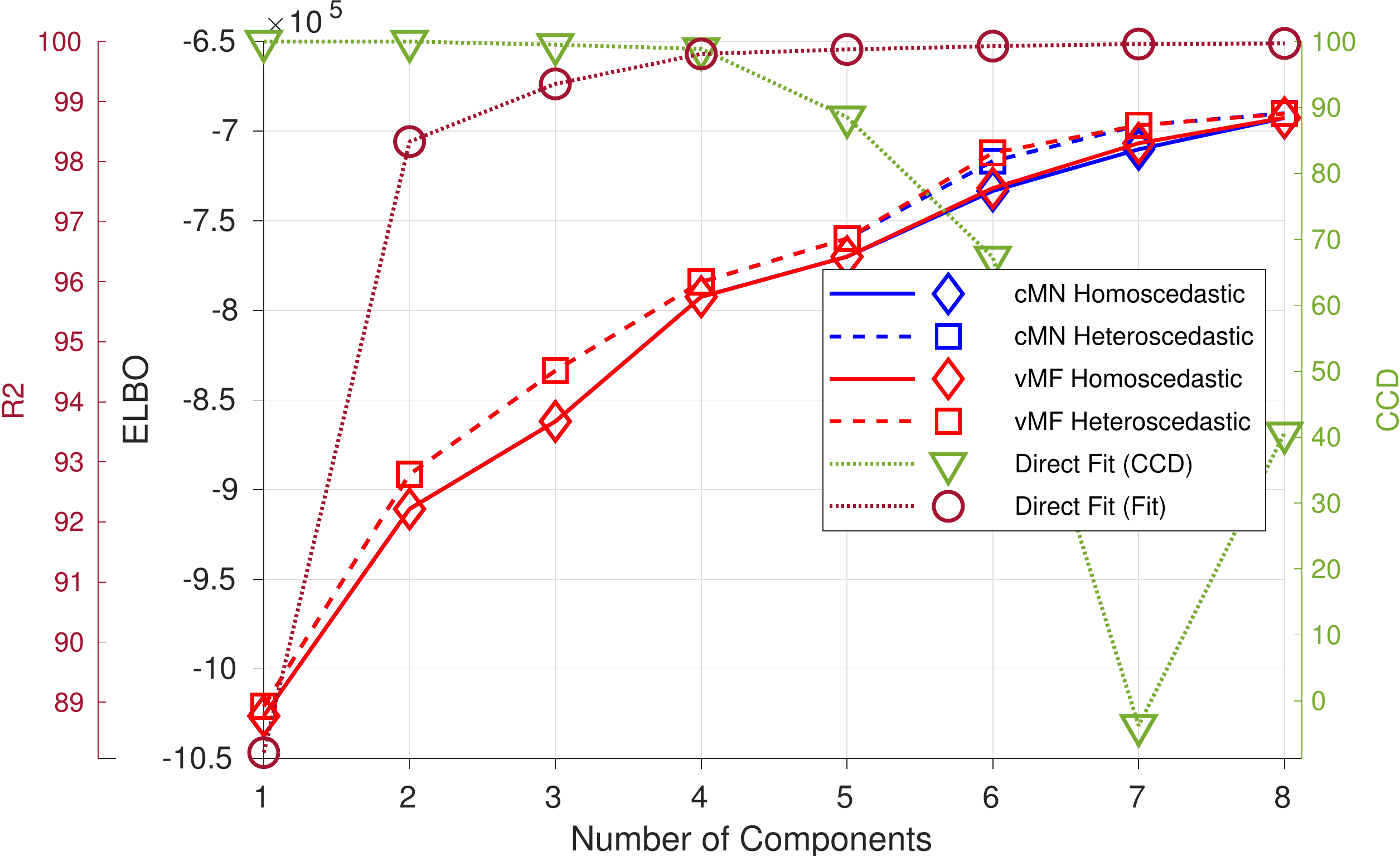}
\label{fig:real_gcms_model_selection}}
\caption{Model selection using the conventional PARAFAC2 and VB-PARAFAC2 models on two real-data sets;  (a) the amino acid flourescence (AAF) and (b) the gas-chromatography mass-spectrometry (GC-MS) data. We report the different model selection criteria R2, CCD, and ELBO.} \label{fig:real_data}
\end{figure*}

\begin{figure*}[!ht]
\centering \includegraphics[width=\linewidth]{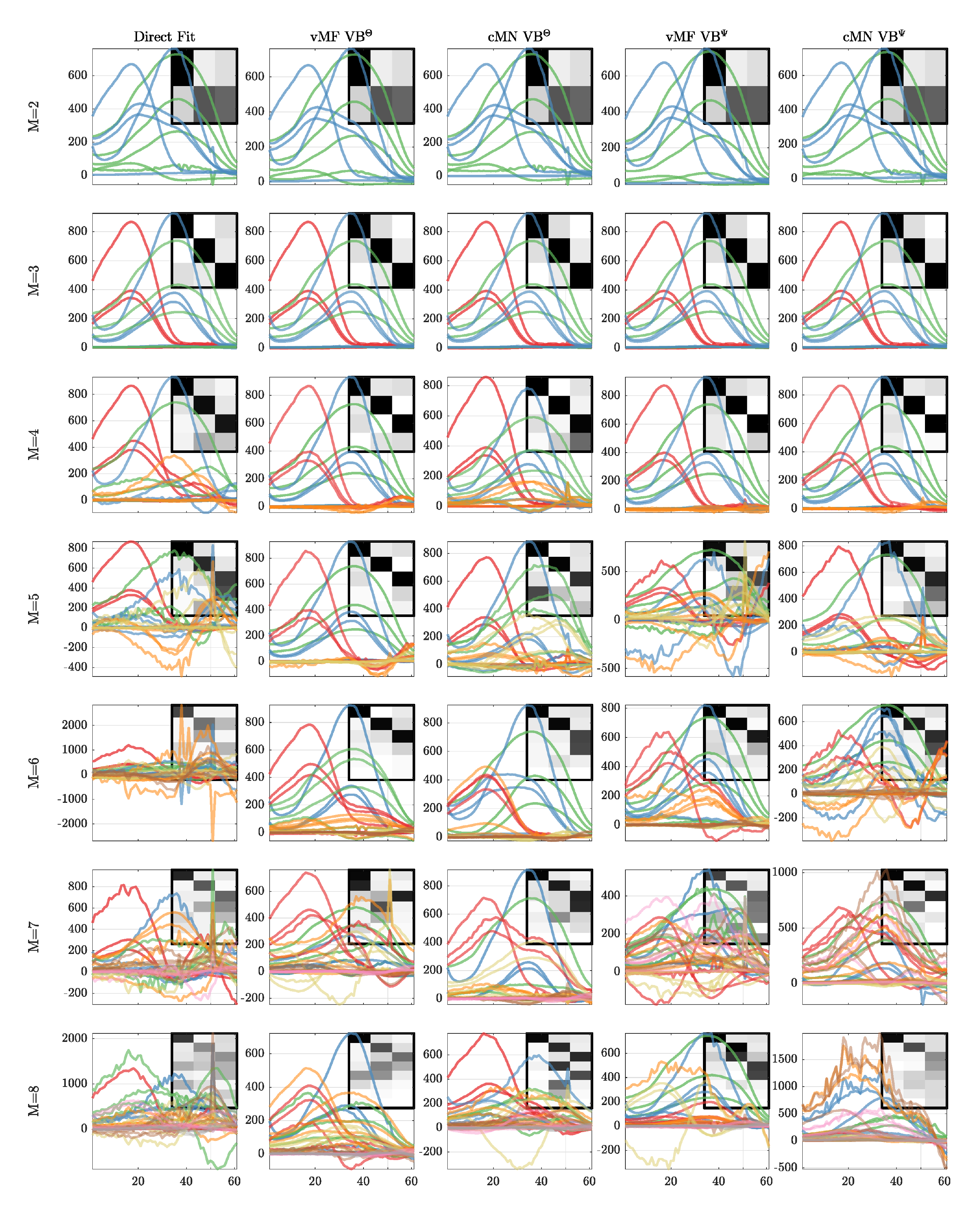}
\caption{The excitation loadings of the amino acid flourescence (AAF) data given by the conventional PARAFAC2 and VB-PARAFAC2 models. From top to bottom the reconstructions uses 2 to 9 components. The heatmap in the background visualizes the correlation the identified components of each model with the components of a conventional PARAFAC2 model with 3 components (ground-truth).} \label{fig:amino_loadings}
\end{figure*}

\begin{figure*}[!ht]
         \centering \includegraphics[width=\linewidth]{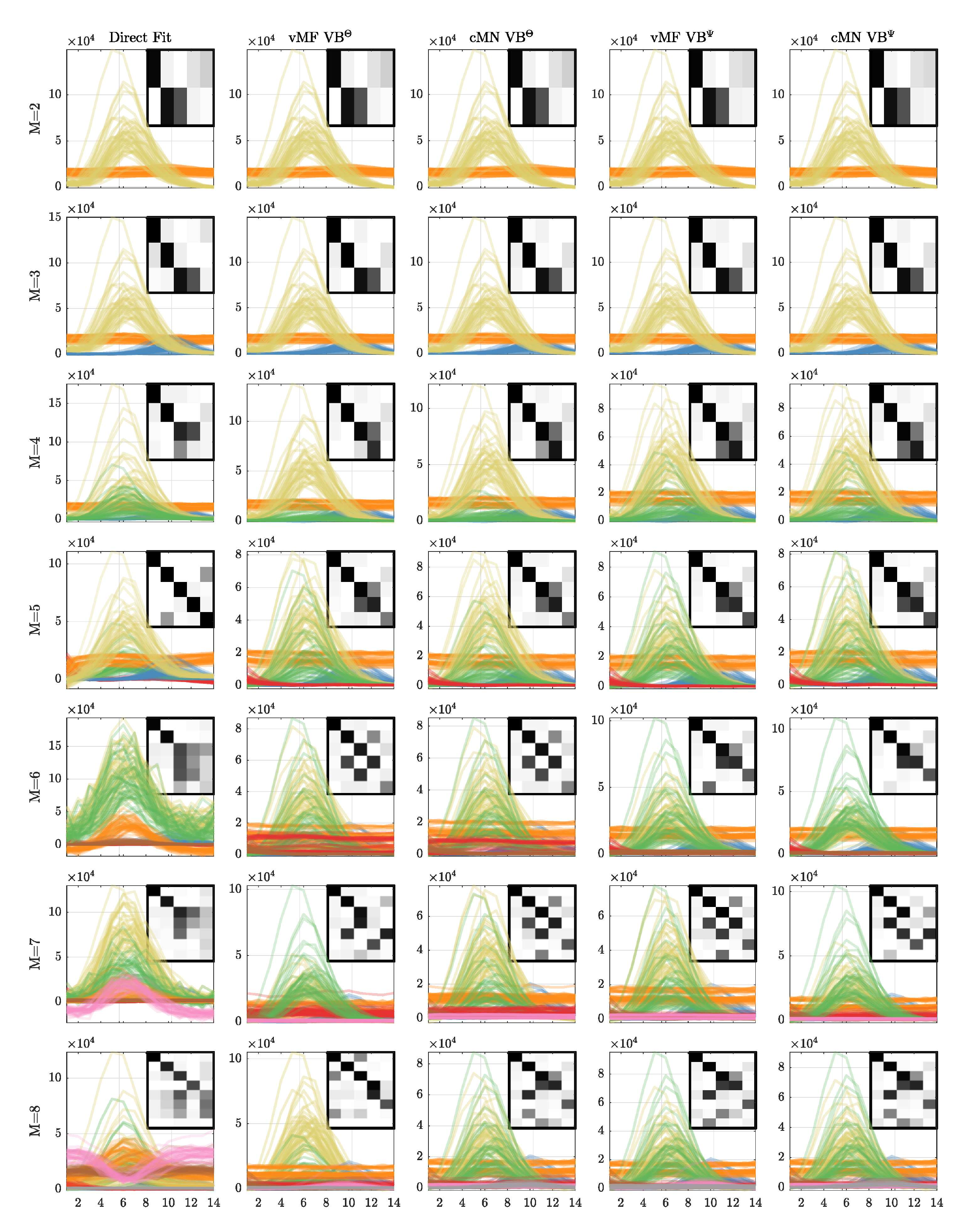}
\caption{The elution profiles of the gas-chromatography mass-spectrometry (GC-MS) data given by the conventional PARAFAC2 and VB-PARAFAC2 models. From top to bottom the reconstructions uses 2 to 9 components. The heatmap in the background visualizes the correlation the identified components of each model with the components of a conventional PARAFAC2 model with 5 components (expert conclusion).} \label{fig:gcms_loadings}
\end{figure*}

The PARAFAC2 has further been used to analyze gas-chromatography mass-spectrometry (GC-MS) data of wine \cite{amigo2008solving,Kamstrup-Nielsen2013}. We presently considered the publicly available\footnote{Available at \url{http://www.models.life.ku.dk/Wine_GCMS_FTIR}} GC-MS data described in \cite{skov2008multiblock}. The dataset contains 44 samples of wine and we here consider the elution time intervals 368-381 of the unaligned data. The data is thus 44 samples of 14 elution times by 200 mass spectrum scans.  In Figure~\ref{fig:real_data}-\ref{fig:gcms_loadings} we consider the estimated components using the direct fitting algorithm and the proposed VB-PARAFAC2 respectively with homo- and heteroscedastic noise. 

In Figure~\ref{fig:real_data} we report the ELBO using the probabilistic models as well as the R2 and \gls{ccd} using the direct fitting algorithm. For the amino acid fluorescence data we observe that both the R2 and \gls{ccd} strongly suggest that a three component model sufficiently describes the data whereas the ELBO finds no substantial improvements beyond three components (Figure~\ref{fig:real_aaf_model_selection}) as well. Investigating the extracted excitation loadings in Figure~\ref{fig:amino_loadings} we observe that both the probabilistic and direct fitting PARAFAC2 extract similar components when too few or the correct number of components are specified, i.e. $\ncomps\leq3$. However, facing misspecification by having chosen too many components the direct fitting algorithm extracts noisy profiles that incorrectly reflect the underlying three constituents whereas the probabilistic PARAFAC2 well recovers the constituents in the homoscedastic noise case up to the specification of $\ncomps=6$ and heteroscedastic noise setting up to $\ncomps=4$. 

For the GC-MS data R2 and \gls{ccd} point to a four or five component model whereas the ELBO points to adding additional components (cf. Figure~\ref{fig:real_gcms_model_selection}).  Inspecting the extracted components in Figure~\ref{fig:gcms_loadings}, we again observe close agreement between the extracted components using the probabilistic and direct fitting PARAFAC2 when specifying a low number of components ($\ncomps\leq5$).  Furthermore, the estimated elution profiles facing model order misspecification appears less influenced by noise than the elution profiles extracted using the direct fitting algorithm emphasizing that including uncertainty in the modeling influence the extracted components.




\section{Conclusion}
We developed a fully probabilistic PARAFAC2 model and demonstrated how orthogonality can be imposed in the context of variational inference in two different ways, i.e. using the von Mises-Fisher matrix distribution assuming $\e{\bm{Y^\top Y}}=\bm{I}$ as proposed in the context of variational PCA in \cite{Smidl2007} but also using the constrained matrix normal distribution assuming $\e{\bm{Y}^{\top}} \e{\bm{Y}}=\bm{I}$ in which the mean is constrained to the Stiefel manifold. For the latter approach we presently derived a simple closed form solution based on optimizing the lower bound. 

Both VB-PARAFAC2 approaches were able to successfully recover the underlying noise free structure in simulated data both when considering homoscedastic and heteroschedastic noise modeling. However, we found that the specification of orthogonality based on vMF was more robust to noise than the specification based on cMN. In particular, we observed substantial noise robustness when compared to the conventional direct fitting approach both when the correct model order was specified and when overestimating the number of components. 

On a simple amino acid fluorescence data the probabilistic PARAFAC2 framework was able to correctly identify the underlying constituents and demonstrated improved robustness to model misspecification when compared to the conventional direct fitting algorithm. For the gas-chromatography mass-spectrometry data we also observed agreement between the probabilistic and direct fitting PARAFAC2 when specifying up to three components, however, substantial differences in the extracted components were observed when including additional components again showing an influence of modeling uncertainty. Variational methods are known to suffer from issues of underestimating uncertainty and thereby becoming overly confident on estimated parameters. While we observed that the probabilistic PARAFAC2 improved robustness on simulated data, real data poses challenges of mismatch between data and convenient variational modeling assumptions. 

Nevertheless, the proposed probabilistic PARAFAC2 forms an important step in the direction of applying probabilistic approaches to more advanced tensor decomposition approaches and a new direction for handling orthogonality constraints in probabilistic modeling in general using the proposed constrained matrix normal distribution framework that has a simple variational update. In particular, we anticipate that the orthogonality constraints within a probabilistic setting may be useful also for the Tucker decomposition in which orthogonality is typically imposed \cite{kolda2009tensor} as well as for block-term decompositions \cite{de2008decompositions} in which orthogonality may be beneficial to impose within each block or to improve identifiability within the CP decomposition by imposing orthogonality as implemented in the n-way toolbox (\url{http://www.models.life.ku.dk/nwaytoolbox}).

\section*{Acknowledgment}
The authors would like to thank Rasmus Bro for providing some of the data analyzed doing this work.
\ifCLASSOPTIONcaptionsoff
\fi



%



\bibliography{Mendeley_philip.bib,bib_file.bib} 

\begin{thebibliography}{10}
\providecommand{\url}[1]{#1}
\csname url@samestyle\endcsname
\providecommand{\newblock}{\relax}
\providecommand{\bibinfo}[2]{#2}
\providecommand{\BIBentrySTDinterwordspacing}{\spaceskip=0pt\relax}
\providecommand{\BIBentryALTinterwordstretchfactor}{4}
\providecommand{\BIBentryALTinterwordspacing}{\spaceskip=\fontdimen2\font plus
\BIBentryALTinterwordstretchfactor\fontdimen3\font minus
  \fontdimen4\font\relax}
\providecommand{\BIBforeignlanguage}[2]{{%
\expandafter\ifx\csname l@#1\endcsname\relax
\typeout{** WARNING: IEEEtran.bst: No hyphenation pattern has been}%
\typeout{** loaded for the language `#1'. Using the pattern for}%
\typeout{** the default language instead.}%
\else
\language=\csname l@#1\endcsname
\fi
#2}}
\providecommand{\BIBdecl}{\relax}
\BIBdecl

\bibitem{Carroll1970}
J.~D. Carroll and J.~J. Chang, ``{Analysis of individual differences in
  multidimensional scaling via an n-way generalization of "Eckart-Young"
  decomposition},'' \emph{Psychometrika}, vol.~35, no.~3, pp. 283--319, 1970.

\bibitem{Harshman1970}
R.~a. Harshman, ``{Foundations of the PARAFAC procedure: Models and conditions
  for an “explanatory” multimodal factor analysis},'' \emph{UCLA Working
  Papers in Phonetics}, vol.~16, no.~10, pp. 1-- 84, 1970.

\bibitem{bro1997parafac}
R.~Bro, ``Parafac. tutorial and applications,'' \emph{Chemometrics and
  intelligent laboratory systems}, vol.~38, no.~2, pp. 149--171, 1997.

\bibitem{Appellof1981-na}
C.~J. Appellof and E.~R. Davidson, ``Strategies for analyzing data from video
  fluorometric monitoring of liquid chromatographic effluents,''
  \emph{Analytical chemistry}, vol.~53, no.~13, pp. 2053--2056, 1981.

\bibitem{kolda2009tensor}
T.~G. Kolda and B.~W. Bader, ``Tensor decompositions and applications,''
  \emph{SIAM review}, vol.~51, no.~3, pp. 455--500, 2009.

\bibitem{morup2011applications}
M.~M{\o}rup, ``Applications of tensor (multiway array) factorizations and
  decompositions in data mining,'' \emph{Wiley Interdisciplinary Reviews: Data
  Mining and Knowledge Discovery}, vol.~1, no.~1, pp. 24--40, 2011.

\bibitem{tucker1966some}
L.~R. Tucker, ``Some mathematical notes on three-mode factor analysis,''
  \emph{Psychometrika}, vol.~31, no.~3, pp. 279--311, 1966.

\bibitem{hitchcock1927expression}
F.~L. Hitchcock, ``The expression of a tensor or a polyadic as a sum of
  products,'' \emph{Studies in Applied Mathematics}, vol.~6, no. 1-4, pp.
  164--189, 1927.

\bibitem{bro1999parafac2}
R.~Bro, C.~A. Andersson, and H.~A. Kiers, ``Parafac2-part ii. modeling
  chromatographic data with retention time shifts,'' \emph{Journal of
  Chemometrics}, vol.~13, no. 3-4, pp. 295--309, 1999.

\bibitem{harshman1996uniqueness}
R.~A. Harshman and M.~E. Lundy, ``Uniqueness proof for a family of models
  sharing features of tucker's three-mode factor analysis and
  parafac/candecomp,'' \emph{Psychometrika}, vol.~61, no.~1, pp. 133--154,
  1996.

\bibitem{ten1996some}
J.~M. ten Berge and H.~A. Kiers, ``Some uniqueness results for parafac2,''
  \emph{Psychometrika}, vol.~61, no.~1, pp. 123--132, 1996.

\bibitem{kiers1999parafac2}
H.~A. Kiers, J.~M. Ten~Berge, and R.~Bro, ``Parafac2-part i. a direct fitting
  algorithm for the parafac2 model,'' \emph{Journal of Chemometrics}, vol.~13,
  no. 3-4, pp. 275--294, 1999.

\bibitem{wise2001application}
B.~M. Wise, N.~B. Gallagher, and E.~B. Martin, ``Application of parafac2 to
  fault detection and diagnosis in semiconductor etch,'' \emph{Journal of
  chemometrics}, vol.~15, no.~4, pp. 285--298, 2001.

\bibitem{weis2010multi}
M.~Weis, D.~Jannek, F.~Roemer, T.~Guenther, M.~Haardt, and P.~Husar,
  ``Multi-dimensional parafac2 component analysis of multi-channel eeg data
  including temporal tracking,'' in \emph{Engineering in Medicine and Biology
  Society (EMBC), 2010 Annual International Conference of the IEEE}.\hskip 1em
  plus 0.5em minus 0.4em\relax IEEE, 2010, pp. 5375--5378.

\bibitem{madsen2016quantifying}
K.~H. Madsen, N.~W. Churchill, and M.~M{\o}rup, ``Quantifying functional
  connectivity in multi-subject fmri data using component models,'' \emph{Human
  Brain Mapping}, 2016.

\bibitem{chew2007cross}
P.~A. Chew, B.~W. Bader, T.~G. Kolda, and A.~Abdelali, ``Cross-language
  information retrieval using parafac2,'' in \emph{Proceedings of the 13th ACM
  SIGKDD international conference on Knowledge discovery and data
  mining}.\hskip 1em plus 0.5em minus 0.4em\relax ACM, 2007, pp. 143--152.

\bibitem{panagakis2011automatic}
Y.~Panagakis and C.~Kotropoulos, ``Automatic music tagging via parafac2,'' in
  \emph{acoustics, speech and signal processing (ICASSP), 2011 IEEE
  international conference on}.\hskip 1em plus 0.5em minus 0.4em\relax IEEE,
  2011, pp. 481--484.

\bibitem{pantraki2015automatic}
E.~Pantraki and C.~Kotropoulos, ``Automatic image tagging and recommendation
  via parafac2,'' in \emph{Machine Learning for Signal Processing (MLSP), 2015
  IEEE 25th International Workshop on}.\hskip 1em plus 0.5em minus 0.4em\relax
  IEEE, 2015, pp. 1--6.

\bibitem{Tian2018-ah}
K.~Tian, L.~Wu, S.~Min, and R.~Bro, ``\BIBforeignlanguage{en}{Geometric search:
  A new approach for fitting {PARAFAC2} models on {GC-MS} data},''
  \emph{\BIBforeignlanguage{en}{Talanta}}, vol. 185, pp. 378--386, Aug. 2018.

\bibitem{Perros2017-ti}
I.~Perros, E.~E. Papalexakis, F.~Wang, R.~Vuduc, E.~Searles, M.~Thompson, and
  J.~Sun, ``{SPARTan}: Scalable {PARAFAC2} for large \& sparse data,'' Mar.
  2017.

\bibitem{Cohen2018-kh}
J.~E. Cohen and R.~Bro, ``Nonnegative {PARAFAC2}: a flexible coupling
  approach,'' Feb. 2018.

\bibitem{porteous2008multi}
I.~Porteous, E.~Bart, and M.~Welling, ``Multi-hdp: A non parametric bayesian
  model for tensor factorization.'' in \emph{Aaai}, vol.~8, 2008, pp.
  1487--1490.

\bibitem{sheng2012probabilistic}
G.~Sheng, L.~Denoyer, P.~Gallinari, and G.~Jun, ``Probabilistic latent tensor
  factorization model for link pattern prediction in multi-relational
  networks,'' \emph{The Journal of China Universities of Posts and
  Telecommunications}, vol.~19, pp. 172--181, 2012.

\bibitem{bhattacharya2011sparse}
A.~Bhattacharya, D.~B. Dunson \emph{et~al.}, ``Sparse bayesian infinite factor
  models,'' \emph{Biometrika}, vol.~98, no.~2, p. 291, 2011.

\bibitem{shan2011probabilistic}
H.~Shan, A.~Banerjee, and R.~Natarajan, ``Probabilistic tensor factorization
  for tensor completion,'' 2011.

\bibitem{Ermis2014}
\BIBentryALTinterwordspacing
B.~Ermis, Y.~K. Yılmaz, a.~T. Cemgil, and E.~Acar, ``{Variational Inference
  For Probabilistic Latent Tensor Factorization with KL Divergence},'' 2014.
  [Online]. Available: \url{http://arxiv.org/abs/1409.8083}
\BIBentrySTDinterwordspacing

\bibitem{xu2012infinite}
Z.~Xu, F.~Yan, and A.~Qi, ``Infinite tucker decomposition: Nonparametric
  bayesian models for multiway data analysis,'' in \emph{Proceedings of the
  29th International Conference on Machine Learning (ICML-12)}, 2012, pp.
  1023--1030.

\bibitem{zhao2015bayesian}
Q.~Zhao, L.~Zhang, and A.~Cichocki, ``Bayesian cp factorization of incomplete
  tensors with automatic rank determination,'' \emph{IEEE transactions on
  pattern analysis and machine intelligence}, vol.~37, no.~9, pp. 1751--1763,
  2015.

\bibitem{hore2016tensor}
V.~Hore, A.~Vi{\~n}uela, A.~Buil, J.~Knight, M.~I. McCarthy, K.~Small, and
  J.~Marchini, ``Tensor decomposition for multiple-tissue gene expression
  experiments,'' \emph{Nature Genetics}, vol.~48, no.~9, pp. 1094--1100, 2016.

\bibitem{beliveau2016spparafac}
V.~Beliveau, G.~Papoutsakis, J.~L. Hinrich, , and M.~Mørup, ``Sparse
  probabilistic parallel factor analysis for the modeling of pet and task-fmri
  data,'' in \emph{Bayesian and Graphiphal Models for Biomedical
  Imaging}.\hskip 1em plus 0.5em minus 0.4em\relax MICCAI, 2016.

\bibitem{Schmidt2009-xs}
M.~N. Schmidt and S.~Mohamed, ``Probabilistic non-negative tensor factorization
  using markov chain monte carlo,'' in \emph{2009 17th European Signal
  Processing Conference}, Aug. 2009, pp. 1918--1922.

\bibitem{xu2015bayesian}
Z.~Xu, F.~Yan, and Y.~Qi, ``Bayesian nonparametric models for multiway data
  analysis,'' \emph{IEEE transactions on pattern analysis and machine
  intelligence}, vol.~37, no.~2, pp. 475--487, 2015.

\bibitem{zhao2016bayesian}
Q.~Zhao, G.~Zhou, L.~Zhang, A.~Cichocki, and S.-I. Amari, ``Bayesian robust
  tensor factorization for incomplete multiway data,'' \emph{IEEE transactions
  on neural networks and learning systems}, vol.~27, no.~4, pp. 736--748, 2016.

\bibitem{hayashi2012exponential}
K.~Hayashi, T.~Takenouchi, T.~Shibata, Y.~Kamiya, D.~Kato, K.~Kunieda,
  K.~Yamada, and K.~Ikeda, ``Exponential family tensor factorization: an online
  extension and applications,'' \emph{Knowledge and information systems},
  vol.~33, no.~1, pp. 57--88, 2012.

\bibitem{cheng2017probabilistic}
L.~Cheng, Y.-C. Wu, and H.~V. Poor, ``Probabilistic tensor canonical polyadic
  decomposition with orthogonal factors,'' \emph{IEEE Transactions on Signal
  Processing}, vol.~65, pp. 663--676, 2017.

\bibitem{Bishop1999}
\BIBentryALTinterwordspacing
C.~M. Bishop, ``{Variational principal components},'' \emph{9th International
  Conference on Artificial Neural Networks ICANN 99}, vol. 1999, no. 470, pp.
  509--514, 1999. [Online]. Available:
  \url{http://link.aip.org/link/IEECPS/v1999/iCP470/p509/s1{\&}Agg=doi}
\BIBentrySTDinterwordspacing

\bibitem{Harshman1972}
\BIBentryALTinterwordspacing
R.~A. Harshman, ``{PARAFAC2: Mathematical and technical notes},'' \emph{UCLA
  Working Papers in Phonetics}, vol.~22, no.~10, pp. 30--44, 1972. [Online].
  Available:
  \url{http://www.bibsonomy.org/bibtex/2a964ff885ba59d4c7be518a3914f737a/threemode}
\BIBentrySTDinterwordspacing

\bibitem{Green1952}
B.~F. Green, ``{The Orthogonal Approximation of An Oblique Staructre in Factor
  Analysis},'' \emph{Psychometrika}, vol.~17, no.~4, pp. 429--440, 1952.

\bibitem{Bro1997-li}
R.~Bro, ``{PARAFAC}. tutorial and applications,'' \emph{Chemometrics Intellig.
  Lab. Syst.}, vol.~38, no.~2, pp. 149--171, 1997.

\bibitem{bro2003new}
R.~Bro and H.~A. Kiers, ``A new efficient method for determining the number of
  components in parafac models,'' \emph{Journal of chemometrics}, vol.~17,
  no.~5, pp. 274--286, 2003.

\bibitem{Kamstrup-Nielsen2013}
M.~H. Kamstrup-Nielsen, L.~G. Johnsen, and R.~Bro, ``{Core consistency
  diagnostic in PARAFAC2},'' \emph{Journal of Chemometrics}, vol.~27, no.~5,
  pp. 99--105, 2013.

\bibitem{Kamstrup-Nielsen2013-wb}
------, ``Core consistency diagnostic in {PARAFAC2},'' \emph{J. Chemom.},
  vol.~27, no.~5, pp. 99--105, 2013.

\bibitem{attias1999variational}
H.~Attias \emph{et~al.}, ``A variational baysian framework for graphical
  models.'' in \emph{NIPS}, vol.~12, 1999.

\bibitem{Bishop2006}
\BIBentryALTinterwordspacing
C.~M. Bishop, \emph{{Pattern Recognition and Machine Learning}}, 2006.
  [Online]. Available:
  \url{http://www.library.wisc.edu/selectedtocs/bg0137.pdf}
\BIBentrySTDinterwordspacing

\bibitem{Blei2016}
\BIBentryALTinterwordspacing
D.~M. Blei, A.~Kucukelbir, and J.~D. McAuliffe, ``{Variational Inference: A
  Review for Statisticians},'' pp. 1--33, 2016. [Online]. Available:
  \url{http://arxiv.org/abs/1601.00670}
\BIBentrySTDinterwordspacing

\bibitem{Smidl2007}
V.~{\v{S}}m{\'{i}}dl and A.~Quinn, ``{On Bayesian principal component
  analysis},'' \emph{Computational Statistics {\&} Data Analysis}, vol.~51,
  no.~9, pp. 4101--4123, 2007.

\bibitem{bro1998multi}
R.~Bro, ``Multi-way analysis in the food industry: models, algorithms, and
  applications,'' Ph.D. dissertation, K{\o}benhavns Universitet, Det
  Biovidenskabelige Fakultet for F{\o}devarer, Veterin{\ae}rmedicin, 1998.

\bibitem{gillis2008nonnegative}
N.~Gillis and F.~Glineur, ``Nonnegative factorization and the maximum edge
  biclique problem,'' \emph{arXiv preprint arXiv:0810.4225}, 2008.

\bibitem{nielsen2014non}
S.~F.~V. Nielsen and M.~M{\o}rup, ``Non-negative tensor factorization with
  missing data for the modeling of gene expressions in the human brain,'' in
  \emph{Machine Learning for Signal Processing (MLSP), 2014 IEEE International
  Workshop on}.\hskip 1em plus 0.5em minus 0.4em\relax IEEE, 2014, pp. 1--6.

\bibitem{khatri1977mises}
C.~Khatri and K.~Mardia, ``The von mises-fisher matrix distribution in
  orientation statistics,'' \emph{Journal of the Royal Statistical Society.
  Series B (Methodological)}, pp. 95--106, 1977.

\bibitem{bishop2006pattern}
C.~M. Bishop, \emph{Pattern recognition and machine learning}.\hskip 1em plus
  0.5em minus 0.4em\relax springer, 2006.

\bibitem{Kiers1999}
H.~A.~L. Kiers, J.~M.~F. Ten~Berge, and R.~Bro, ``{PARAFAC2 — Part I . A
  Direct Fitting Algorithm for the PARAFAC2 Model},'' \emph{J. Chemom.},
  vol.~13, pp. 275--294, 1999.

\bibitem{kiers1998three}
H.~A. Kiers, ``A three-step algorithm for candecomp/parafac analysis of large
  data sets with multicollinearity,'' \emph{Journal of Chemometrics}, vol.~12,
  no.~3, pp. 155--171, 1998.

\bibitem{amigo2008solving}
J.~M. Amigo, T.~Skov, R.~Bro, J.~Coello, and S.~Maspoch, ``Solving gc-ms
  problems with parafac2,'' \emph{TrAC Trends in Analytical Chemistry},
  vol.~27, no.~8, pp. 714--725, 2008.

\bibitem{skov2008multiblock}
T.~Skov, D.~Ballabio, and R.~Bro, ``Multiblock variance partitioning: A new
  approach for comparing variation in multiple data blocks,'' \emph{analytica
  chimica acta}, vol. 615, no.~1, pp. 18--29, 2008.

\bibitem{de2008decompositions}
L.~De~Lathauwer, ``Decompositions of a higher-order tensor in block
  terms—part ii: Definitions and uniqueness,'' \emph{SIAM Journal on Matrix
  Analysis and Applications}, vol.~30, no.~3, pp. 1033--1066, 2008.

\end{thebibliography}
\bibliographystyle{IEEEtran}
%

\begin{IEEEbiography}[{\includegraphics[width=1in,height=1.25in,clip,keepaspectratio]{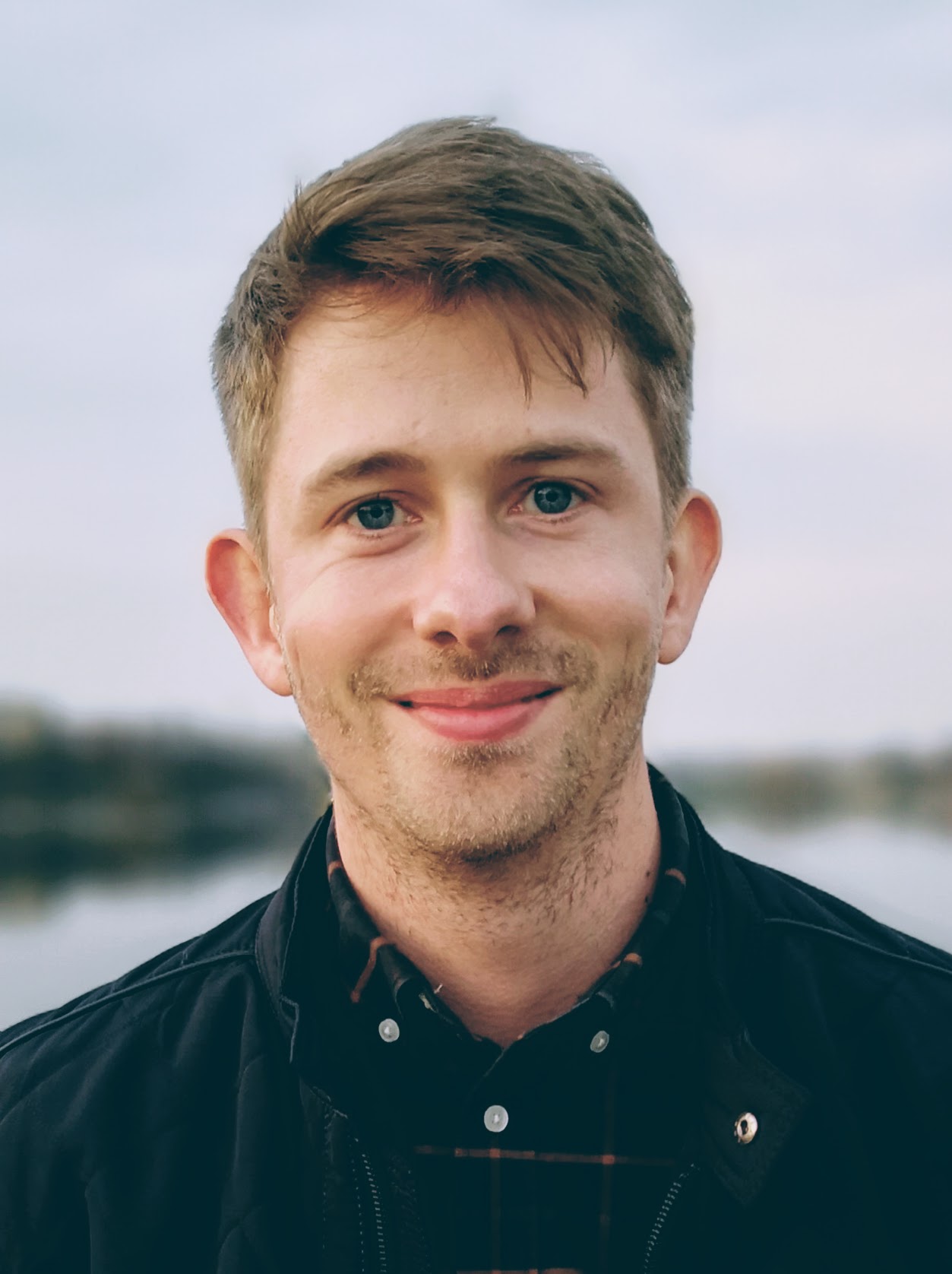}}]{Philip J. H. J\o rgensen}
received his B.Sc. and M.Sc. degrees in mathematical modelling and computation from the Technical University of Denmark (DTU) in 2014 and 2016, respectively. Currently, he is a PhD student at the intersection of the statistics and data analysis section and cognitive systems section at DTU. His research focus includes probabilistic methods and lifelong machine learning.
\end{IEEEbiography}
\begin{IEEEbiography}[{\includegraphics[width=1in,height=1.25in,clip,keepaspectratio]{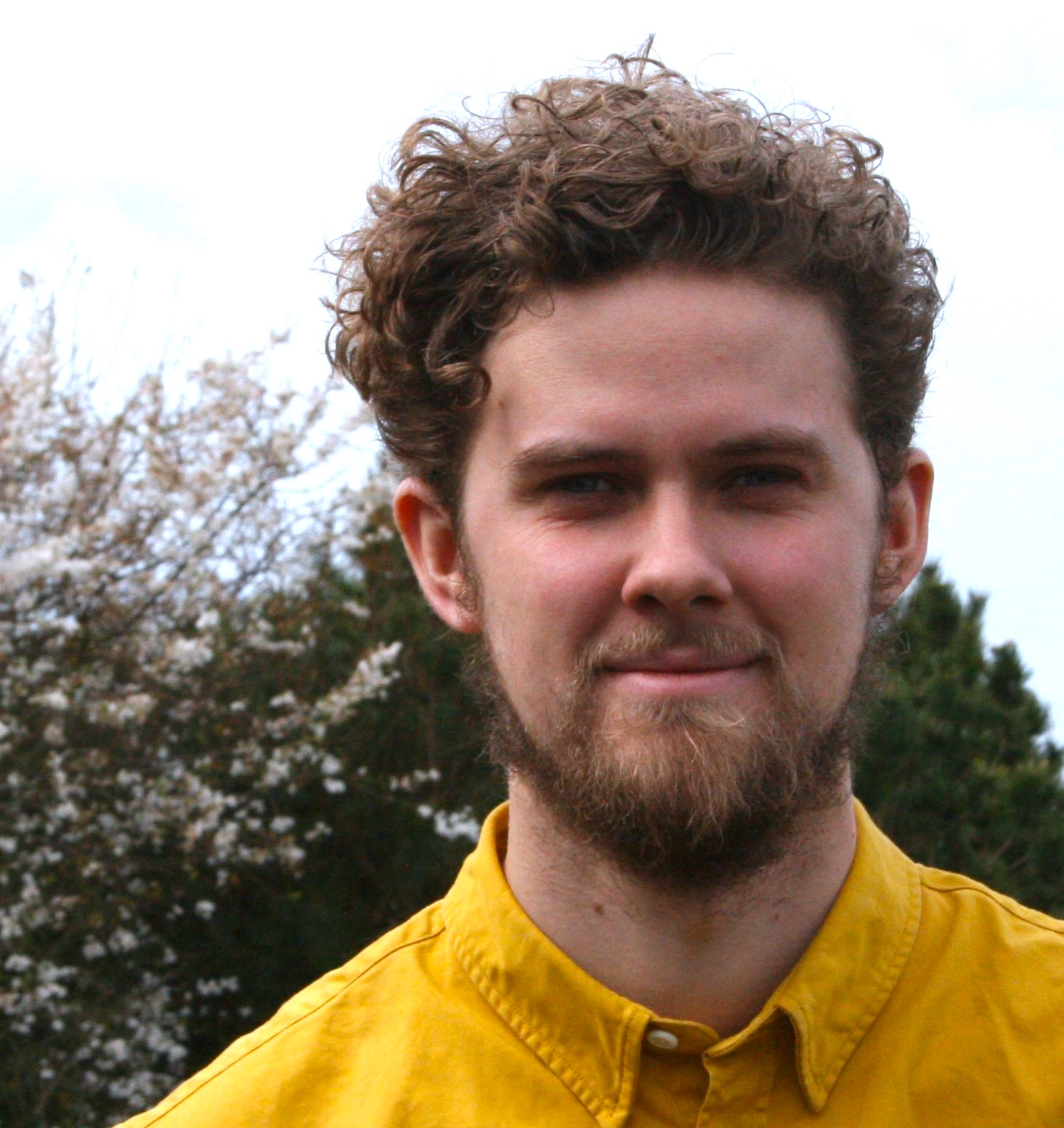}}]{S\o ren F. V. Nielsen}
S\o ren F. V. Nielsen received a B.Sc. and M.Sc. degree in mathematical modeling and engineering from the Technical University of Denmark (DTU) 2013 and 2015, respectively. He is now currently pursuing a Ph.d. in machine learning also at DTU. His research interest are within the field of Bayesian machine learning applied to neuroimaging, including modeling dynamic functional connectivity. 
\end{IEEEbiography}
\begin{IEEEbiography}[{\includegraphics[width=1in,height=1.25in,clip,keepaspectratio]{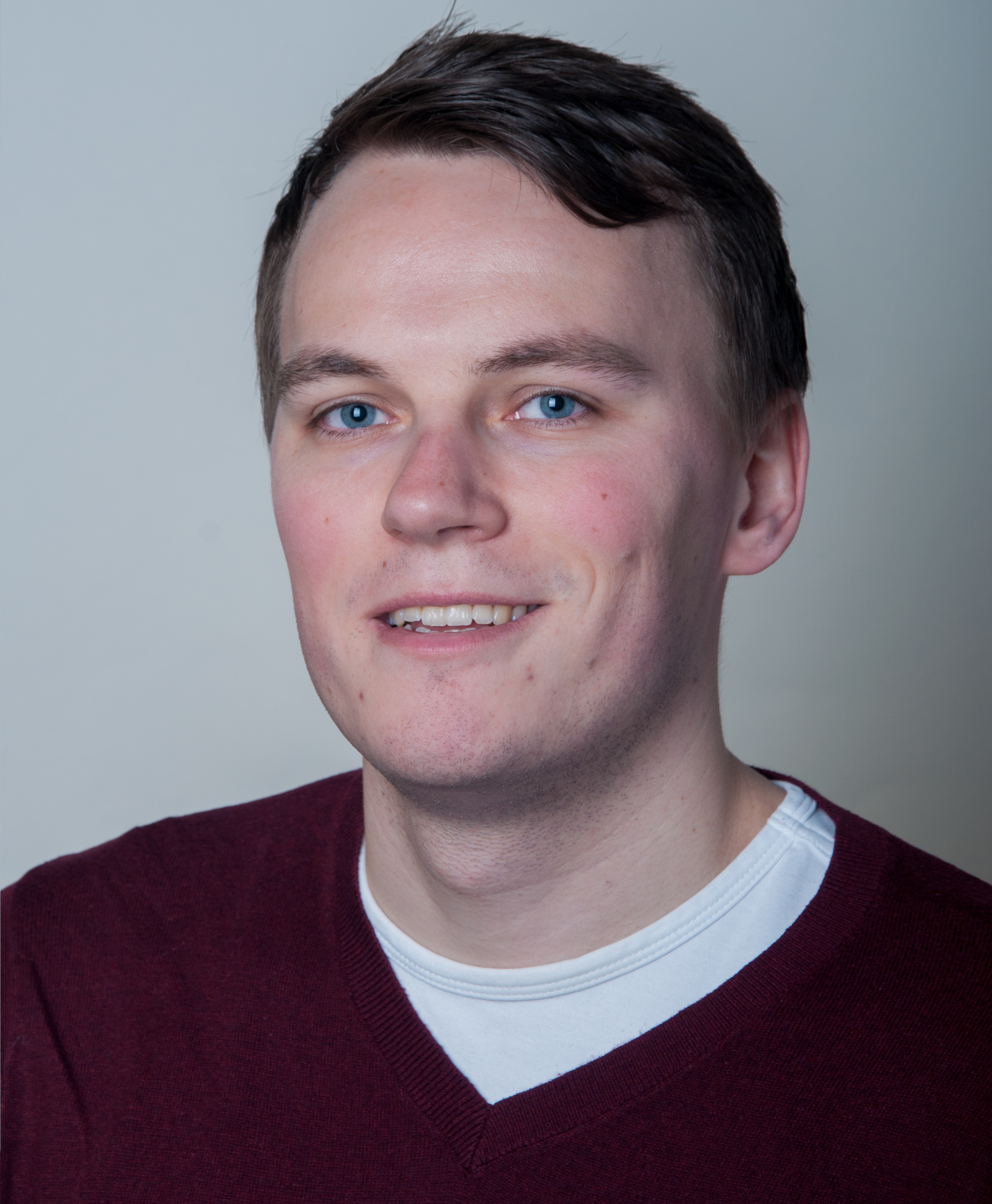}}]{Jesper L{\o}ve Hinrich}
received the B.Sc. degree in mathematics and technology and the M.Sc. degree in mathematical modelling and computation from the Technical University of Denmark in 2013 and 2016, respectively.

He is currently a Ph.D. student at the Technical University of Denmark, DTU Compute, Department for Cognitive Systems. His primary interests are in probabilistic multi-way modeling, machine learning and biomedical imaging.
\end{IEEEbiography}
\begin{IEEEbiography}[{\includegraphics[width=1in,height=1.25in,clip,keepaspectratio]{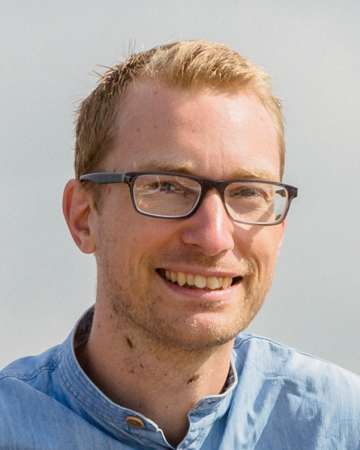}}]{Mikkel N. Schmidt} is Associate Professor
at DTU Compute, Technical University of Denmark. He is interested in probabilistic modeling and statistical machine learning with applications in both science and industry. His primary focus is the development of computational procedures for inference and validation.
\end{IEEEbiography}
\begin{IEEEbiography}[{\includegraphics[width=1in,height=1.25in,clip,keepaspectratio]{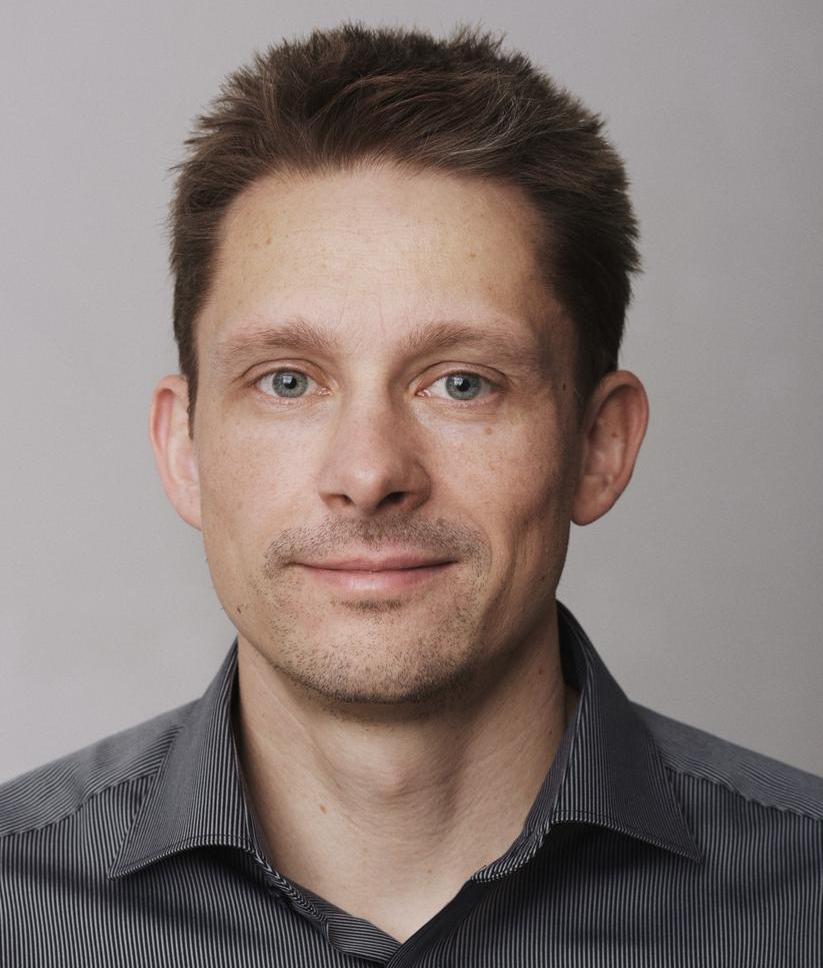}}]{Kristoffer H. Madsen}
Kristoffer H. Madsen received his Ph.D. degree from the Technical University of Denmark (DTU), he is currently heading the computational neuroimaging group at the Danish Research Centre for Magnetic Resonance and has an academic appointment as associate professor at DTU Compute. His primary research focus is on statistical machine learning for functional neuroimaging applications.
\end{IEEEbiography}
\begin{IEEEbiography}[{\includegraphics[width=1in,height=1.25in,clip,keepaspectratio]{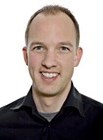}}]{Morten M\o rup}
Morten M\o rup (mmor@dtu.dk) received his M.S. and Ph.D. degrees in applied mathematics at the Technical University of Denmark and he is currently associate professor at the Section for Cognitive Systems at DTU Compute, Technical University of Denmark. He has been associate editor of IEEE Transactions on Signal Processing. His research interests include machine learning, neuroimaging, and complex network modeling.
\end{IEEEbiography}







\newpage

\renewcommand\thesection{\Alph{section}}
\renewcommand\thesection{\Alph{section}}
\renewcommand\thesubsection{\Alph{subsection}}

\begin{onecolumn}
\begin{center}
\textbf{\large Supplemental Material}
\end{center}
\setcounter{equation}{0}
\setcounter{section}{0}
\setcounter{figure}{0}
\setcounter{table}{0}
\setcounter{page}{1}
\makeatletter
\renewcommand{\theequation}{S\arabic{equation}}
\renewcommand{\thefigure}{S\arabic{figure}}

\section{Software}
A MATLAB implementation of the probabilistic PARAFAC2 model was used to run all experiments and generate the results in the paper. The source code is made available on GitHub here (\url{https://github.com/philipjhj/VBParafac2}). You find a guide on setup and usage on GitHub.


\section{Deriving the Variational Inference}
In the following we derive the most important expressions used to identify the update rules of the model parameters.  Below is a overview of the used notation.

\nomenclature[0]{$\vara$}{Matrix (bold face uppercase)}
\nomenclature[0.1]{$\transpose\vara$}{Transpose of matrix $\vara$}
\nomenclature[0.2]{$\varavec$}{i'th row of matrix $\vara$ (bold face lowercase)}
\nomenclature[0.5]{$\varacol$}{i'th column of matrix $\vara$ (bold face lowercase)}
\nomenclature[1]{$\fullx$}{Tensor / Multi-way array (calligraphy)}
\nomenclature[2]{$\varx$}{k'th frontal slice (matrix) of tensor $\fullx$}
\nomenclature[3]{$\varxvec$}{i'th row of the k'th frontal slice $\varx$ of tensor $\fullx$}
\nomenclature[4]{$\pDist\cdot$}{Probability density function (pdf)}
\nomenclature[5]{$\qDist\cdot$}{Variational distribution}
\nomenclature[6]{$\e\cdot$}{Expectation}
\nomenclature[7]{$\ej{z}{\cdot}$}{Expectation with respect to all variables but $z$}
\nomenclature[8]{$\entropy{\cdot}$}{Entropy}
\nomenclature[9]{$\cn{x}$}{Constant term(s) $x$ irrelevant for the optimization problem at hand}
\nomenclature[9]{$\hadamard$}{The Hadamard product (element-wise)}
\printnomenclature

\subsection{The Evidence Lower Bound (ELBO)}
An expansion of the \gls{elbo} is shown here:
\begin{align*}
	\text{ELBO}(\qDist{ \varparameters })=&\elog{  p(\fullx,\varparameters)}-\elog{  \qDist{ \varparameters } } \\
=&\e{\log \pDist{\fullx,\allparameters}}-\elog{ \qDist{\allparameters}} \\
	=&\elog{  \pDistCon{\fullx}{\vara,\varc,\varf,\fullp,\varsigma}\pDist{\vara}\pDistCon{\varc}{\varalpha}\pDist{\varalpha}\pDist{\varf}\pDist{\fullp}\pDist{\varsigma} } \\
	&-\elog{\qDist{\vara}\qDist{\varc}\qDist\varf\qDist\varp\qDist\varsigma\qDist\varalpha}\\
	=&\elog{\pDistCon{\fullx}{\vara,\varc,\varf,\fullp,\varsigma}}+\elog{\pDist{\vara}}+\elog{\pDistCon{\varc}{\varalpha}}\\
	&+\elog{\pDist{\varalpha}}+\elog{\pDist{\varf}}+\elog{\pDist{\fullp}}+\elog{\pDist{\varsigma}}  \\
	&-\elog{\qDist{\vara}}-\elog{\qDist{\varc}}-\elog{\qDist{\varf}}-\elog{\qDist{\fullp}}\\
	&-\elog{\qDist{\varsigma}}-\elog{\qDist{\varalpha}} \\
	=&\elog{\pDistCon{\fullx}{\vara,\varc,\varf,\fullp,\varsigma}}+\elog{\pDist{\vara}}+\elog{\pDistCon{\varc}{\varalpha}}\\
	&+\elog{\pDist{\varalpha}}+\elog{\pDist{\varf}}+\elog{\pDist{\fullp}}+\elog{\pDist{\varsigma}}  \\
	&+\entropy{\qDist{\vara}}+\entropy{\qDist{\varc}}+\entropy{\qDist{\varf}}+\entropy{\qDist{\fullp}}\\
	&+\entropy{\qDist{\varsigma}}+\entropy{\qDist{\varalpha}} \\
\end{align*}

How to derive each of these terms is shown in the following.

\subsection{Standard Moment Matching}
\label{sec:mmatch}
As the formulation of the probabilistic PARAFAC2 model consists of the multivariate normal and gamma distribution we expand the logarithm of their general expressions below. This will serve as a reference for identifying the parameters of the variational distribution when reading the derivations of the update rules.

\subsubsection{Multivariate Normal Distribution}
Deriving the log density function of the multivariate normal distribution amounts to:
\begin{align*}
f(x_1,\hdots,x_k) &\sim \normdist{X}	\\
f(x_1,\hdots,x_k) &= (2\pi)^{-\frac{k}{2}}(|\Sigmasubbf{X}|)^{-\frac{1}{2}}\exp(-\frac{1}{2}\transpose{(\mathbf{X}-\musubbf{X})}\inverse{\Sigmasubbf{X}} (\mathbf{X}-\musubbf{X})) \\
\Rightarrow \ln f(x_1,\hdots,x_k) &= \ln\left[(2\pi)^{-\frac{k}{2}}(|\Sigmasubbf{X}|)^{-\frac{1}{2}}\exp\left(-\frac{1}{2}\transpose{(\mathbf{X}-\musubbf{X})}\inverse{\Sigmasubbf{X}} (\mathbf{X}-\musubbf{X})\right)\right] \\
 &= -\frac{k}{2}\ln(2\pi)-\frac{1}{2}\ln(|\Sigmasubbf{X}|)-\frac{1}{2}\transpose{(\mathbf{X}-\musubbf{X})}\inverse{\Sigmasubbf{X}} (\mathbf{X}-\musubbf{X}) \\
 &= -\frac{k}{2}\ln(2\pi)-\frac{1}{2}\ln(|\Sigmasubbf{X}|)-\frac{1}{2}\transpose{\mathbf{X}}\inverse{\Sigmasubbf{X}}\mathbf{X}-\frac{1}{2}\transpose{\musubbf{X}}\inverse{\Sigmasubbf{X}}\musubbf{X}+\transpose{\musubbf{X}}\inverse{\Sigmasubbf{X}}\mathbf{X} \\
 &= -\frac{1}{2}\ln(|\Sigmasubbf{X}|)-\frac{1}{2}\transpose{\mathbf{X}}\inverse{\Sigmasubbf{X}}\mathbf{X} -\frac{1}{2}\transpose{\musubbf{X}}\inverse{\Sigmasubbf{X}}\musubbf{X}+\transpose{\musubbf{X}}\inverse{\Sigmasubbf{X}}\mathbf{X}+c \\
\end{align*}
where $c$ is the constant terms with respect to $\varx$ and its parameters.

\subsubsection{Gamma Distribution}
Deriving the log density function of the gamma distribution amounts to:
\begin{align*}
f(x; a,b) =& \frac{1}{\Gamma(a)b^a}x^{a-1}\exp(-xb^{-1}) \\
\Rightarrow \ln f(x; a,b) =& \ln\left[ \frac{1}{\Gamma(a)b^a}x^{a-1}\exp(-xb^{-1})\right]\\
=&  \ln\frac{1}{\Gamma(a)b^a}+(a-1)\ln x-xb^{-1} \\
=& (a-1)\ln x-xb^{-1} +c\\
\end{align*}
where $c$ is the constant terms with respect to $x$.

\subsection{Non-trivial Moment Matching}
To identify the parameters for $\varc$ and $\varf$ non-trivial steps had to be performed.
\subsubsection{The F Matrix}

The variational factor for $\varf$ is defined as: 

\begin{align*}
{\qDist{\varf}} \propto& \exp\elogj{\varf}{ \pDist{\fullx,\varparameters}} \\
\propto& \exp\ej{\varf}{\log \pDistCon{\fullx,\varf}{\vara,\varc,\fullp,\varsigma}}
\end{align*}
where
\begin{align*}
\ej{\varf}{\log \pDistCon{\fullx,\varf}{\vara,\varc,\fullp,\varsigma}}
=& \elogj{\varf}{\pDistCon{\fullx}{\vara,\varc,\varf,\fullp,\varsigma}}+\elogj{\varf}{\pDist{\varf}} \\
=& \sum_k\sum_i\elogj{\varf}{\pDistCon{\varxvec}{\varavec,\vardvec,\varf,\varp,\varsigmavec}}+\sum_m\elogj{\varf}{\pDist{\varfvec}} \\
=& \sum_k\sum_i\ej{\varf}{-\frac{1}{2}(\pADFP)\Identity{M}\varsigmavec\transpose{(\pADFP)} \\
&+\varavec\vardvec\transpose{\varf}\transpose{\varp}\Identity{M}\varsigmavec\transpose{\varxvec}}+\sum_m\ej{\varf}{-\frac{1}{2}\varfvec \Identity{M}\transpose\varfvec}+ c \\
=&-\frac{1}{2}\sum_k\sum_i\ej{\varf}{\varsigmavec(\varavec\vardvec\transpose\varf\transpose{\varp}\varp\varf\vardvec\transpose\varavec)}-\frac{1}{2}\sum_m\varfvec\transpose\varfvec \\
&+\sum_k\sum_i\ej{\varf}{ \varsigmavec\varavec\vardvec\transpose\varf\transpose{\varp}\transpose{\varxvec}}+ c \\
=&-\frac{1}{2}\sum_k\e{\varsigmavec}\sum_i\ej{\varf}{\varavec\vardvec\transpose\varf\transpose{\varp}\varp\varf\vardvec\transpose\varavec}-\frac{1}{2}\sum_m\varfvec\transpose\varfvec\\
&+\sum_k\sum_i\e{\varsigmavec}\ej{\varf}{ \varavec\vardvec\transpose\varf\transpose{\varp}\transpose{\varxvec}}+ c \\
\end{align*}
Again, we reorder the parameters using the trace operator to identify the quadratic term. This time the quadratic term separates into a quadratic and linear part revealing a linear intercomponent dependency.

\begin{align*}
\ej{\varf}{\varavec\vardvec\transpose\varf\transpose{\varp}\varp\varf\vardvec\transpose\varavec} =& \ej{\varf}{\trace{\varavec\vardvec\transpose\varf\transpose{\varp}\varp\varf\vardvec\transpose\varavec}}\\
=& \ej{\varf}{\trace{\varf\vardvec\transpose\varavec\varavec\vardvec\transpose\varf\transpose{\varp}\varp}}\\
=& \trace{\varf\ej{\varf}{\vardvec\transpose\varavec\varavec\vardvec}\transpose\varf\ej{\varf}{\transpose{\varp}\varp}}\\
=& \sum_{mm\prime}(\varf\e{\vardvec\transpose\varavec\varavec\vardvec}\transpose\varf)_{mm\prime}(\e{\transpose{\varp}\varp})_{mm\prime}\\
=& \sum_{mm\prime}\varfvec\e{\vardvec\transpose\varavec\varavec\vardvec}\mathbf{f}_{m\prime\cdot}^\text{T}\e{\mathbf{p}^\text{T}_{\cdot m k}\mathbf{p}_{\cdot m\prime k}}\\
=& \sum_m \varfvec\e{\vardvec\transpose\varavec\varavec\vardvec}\e{\mathbf{p}^\text{T}_{\cdot m k}\mathbf{p}_{\cdot m k}}\mathbf{f}_{m\cdot}^\text{T}\\
 &+ 2\sum_m\sum_{m\prime \setminus m} \varfvec\e{\vardvec\transpose\varavec\varavec\vardvec}\e{\mathbf{p}^\text{T}_{\cdot m k}\mathbf{p}_{\cdot m\prime k}}\mathbf{f}_{m\prime\cdot}^\text{T}\\
\end{align*}

Again, we have to reorder and include the linear terms as before.

\begin{align*}
\sum_k\sum_i\e{\varsigmavec}\ej{\varf}{ \varavec\vardvec\transpose\varf\transpose{\varp}\transpose{\varxvec}} =& \sum_k\sum_i\e{\varsigmavec}\sum_m\ej{\varf}{ \varavec\vardvec\transpose\varfvec(\transpose{\varp})_m\transpose{\varxvec}} \\
=& \sum_k\sum_i\e{\varsigmavec}\sum_m\e{\varavec}\e{\vardvec}\transpose\varfvec\e{(\transpose{\varp})_m}\transpose{\varxvec} \\
=& \sum_k\sum_i\e{\varsigmavec}\sum_m\e{(\transpose{\varp})_m}\transpose{\varxvec}\e{\varavec}\e{\vardvec}\transpose\varfvec\\
=& \sum_k\e{\varsigmavec}\sum_m\e{(\transpose{\varp})_m}(\sum_i\transpose{\varxvec}\e{\varavec})\e{\vardvec}\transpose\varfvec
\end{align*}

Accounting for all terms and matching them to the ones in \reff{sec:mmatch} we arrive at the following update rules for $\varf$.

\begin{flalign}
&\qDist{\varf} = \prod_m \normdistp{\musubbf{\varfvec}}{\Sigmasubbf{\varfvec}} \\
	&\musubbf{\varfvec} = \Sigmasubbf{\varfvec}(\sum_k\e{\varsigmavec}(\e{(\transpose{\varp})_m}\transpose{\varx}\e{\vara}\e{\vardvec}-\sum_i\e{\vardvec\transpose\varavec\varavec\vardvec}\sum_{m\prime \setminus m} \e{\mathbf{p}^\text{T}_{\cdot m k}\mathbf{p}_{\cdot m\prime k}}\mathbf{f}_{m\prime\cdot}^\text{T}))&\\
	& \Sigmasubbf{\varfvec} = \inverse{(\sum_k\e{\varsigmavec}\sum_i\e{\vardvec\transpose\varavec\varavec\vardvec}\e{\mathbf{p}^\text{T}_{\cdot m k}\mathbf{p}_{\cdot m\prime k}}+\Identity{M})}&
\end{flalign}

\subsection{Other Expressions}

\subsubsection{Constrained Matrix Normal Distribution}
The orthogonality constraint in the model can be handled with two formulations. This section concerns the approach where the mean parameters of the variational approximation for $\varp$ is constrained to be orthogonal and the following section describes the solution using the von-Mises Fisher distribution. Instead of using the free form variational updates we optimizes the ELBO with respect to the mean parameters $\Msub{\varp}=\e\varp$ constrained to be orthogonal.

\begin{align*}
\Msub{\varp}=&\underset{\Msub{\varp}}{\text{arg max}} ~\text{ELBO}(\Msub{\varp}) &\text{subject to}~ \Msub{\varp}\transpose{\Msub{\varp}} = \Identity{}  \\
=&\underset{\Msub{\varp}}{\text{arg max}}~\elog{\pDistCon{\fullx}{\vara,\varc,\varf,\fullp,\varsigma}}+\elog{\pDist{\vara}}+\elog{\pDistCon{\varc}{\varalpha}}\\
	&+\elog{\pDist{\varalpha}}+\elog{\pDist{\varf}}+\elog{\pDist{\fullp}}+\elog{\pDist{\varsigma}}  \\
	&+\entropy{\qDist{\vara}}+\entropy{\qDist{\varc}}+\entropy{\qDist{\varf}}+\entropy{\qDist{\fullp}}\\
	&+\entropy{\qDist{\varsigma}}+\entropy{\qDist{\varalpha}} &\text{subject to}~ \Msub{\varp}\transpose{\Msub{\varp}} = \Identity{}  \\
    =&\underset{\Msub{\varp}}{\text{arg max}} ~\elog{\pDistCon{\fullx}{\vara,\varc,\varf,\fullp,\varsigma}} + \cn{1}&\text{subject to}~ \Msub{\varp}\transpose{\Msub{\varp}} = \Identity{}  \\
    =&\underset{\Msub{\varp}}{\text{arg max}} ~-\frac{1}{2}\sum_k\sum_i\e{\varsigmavec(\varavec\vardvec\transpose\varf\transpose{\varp}\varp\varf\vardvec\transpose\varavec)} \\
&+\sum_k\sum_i\e{ \varsigmavec\varavec\vardvec\transpose\varf\transpose{\varp}\transpose{\varxvec}}+\cn{2}
&\text{subject to}~ \Msub{\varp}\transpose{\Msub{\varp}} = \Identity{}  \\
    =&\underset{\Msub{\varp}}{\text{arg max}} ~\sum_k\sum_i\e{ \varsigmavec\varavec\vardvec\transpose\varf\transpose{\varp}\transpose{\varxvec}}+\cn{3}
    &\text{subject to}~ \Msub{\varp}\transpose{\Msub{\varp}} = \Identity{}  \\
    =&\underset{\Msub{\varp}}{\text{arg max}} ~\sum_k\e{ \varsigmavec}\trace{\e\vara\e\vardvec\e{\transpose\varf}\e{\transpose{\varp}}\transpose{\varx}}+\cn{3}
    &\text{subject to}~ \Msub{\varp}\transpose{\Msub{\varp}} = \Identity{}  \\
    =&\underset{\Msub{\varp}}{\text{arg max}} ~\sum_k\e{ \varsigmavec}\trace{\e{\varf}\e\vardvec\e{\transpose{\vara}}\varx\Msub{\varp}}+\cn{3}
    &\text{subject to}~ \Msub{\varp}\transpose{\Msub{\varp}} = \Identity{}  \\
\end{align*}

Only the linear term of the probability density function of the data $\fullx$ depends on $\Msub{\varp}$ since $\Msub{\varp}$ in the quadratic terms is the identity matrix. Except for a scalar the optimization problem reduces to the same one as finding $\varp$ in the alternating least squares algorithm, where one maximizes $\trace{\e{\varf}\e\vardvec\e{\transpose{\vara}}\varx\Msub{\varp}}$ subject to the orthogonality constraint. The solution to this is found by simply applying a SVD as stated in the main text\footnote{The alternating least squares method is described in \emph{Kiers, Henk A. L., Jos M. F. Ten Berge, and Rasmus Bro. 1999. “PARAFAC2 — Part I. A Direct Fitting Algorithm for the PARAFAC2 Model.” Journal of Chemometrics 13: 275–94.} and the solution to the optimization problem was first described in \emph{Green, Bert F. 1952. “The Orthogonal Approximation of An Oblique Staructre in Factor Analysis.” Psychometrika 17 (4): 429–40.}}.

\subsubsection{The C matrix}
The variational factor for $\varc$ is defined as: 
\begin{align*}
{\qDist{\varc}} \propto& \exp\elogj{\varc}{ \pDist{\fullx,\varparameters}} \\
\propto& \exp\ej{\varc}{\log \pDistCon{\fullx,\varc}{\vara,\varf,\fullp,\varsigma,\varalpha}}
\end{align*}
where
\begin{align*}
	\hspace{-50pt}\ej{\varc}{\log \pDistCon{\fullx,\varc}{\vara,\varf,\fullp,\varsigma,\varalpha}}
=& \elogj{\varc}{\pDistCon{\fullx}{\vara,\varc,\varf,\fullp,\varsigma}}+\elogj{\varc}{\pDistCon{\varc}{\varalpha}} \\
=& \sum_k\sum_i\elogj{\varc}{\pDistCon{\varxvec}{\varavec,\vardvec,\varf,\varp,\varsigmavec}}+\sum_k\elogj{\varc}{\pDistCon{\varcvec}{\varalpha}} \\
=& \sum_k\sum_i\ej{\varc}{-\frac{1}{2}(\pADFP)\Identity{M}\varsigmavec\transpose{(\pADFP)} \\
 &+\pADFP\Identity{M}\varsigmavec\transpose{\varxvec}}+\sum_k \ej{\varc}{-\frac{1}{2}\varcvec \diag{\varalpha}\transpose\varcvec}+ c \\
=&-\frac{1}{2}\sum_k\sum_i\e{\varsigmavec}\ej{\varc}{\varavec\vardvec\transpose\varf\transpose{\varp}\varp\varf\vardvec\transpose\varavec} \\
 &+\e{\varsigmavec}\ej{\varc}{\varavec\vardvec\transpose{\varf}\transpose{\varp}\transpose{\varxvec}}-\frac{1}{2}\sum_k \varcvec \diag{\e{\varalpha}}\transpose\varcvec+ c \\
=&-\frac{1}{2}\sum_k\left(\e{\varsigmavec}\left(\sum_i\ej{\varc}{\varavec\vardvec\transpose\varf\transpose{\varp}\varp\varf\vardvec\transpose\varavec}\right)+ \varcvec \Identity{M}\e{\varalpha}\transpose\varcvec\right)\\
&+\sum_k\sum_i\e{\varsigmavec}\ej{\varc}{\varavec\vardvec\transpose{\varf}\transpose{\varp}\transpose{\varxvec}}+ c \\
\end{align*}

Where $\diag{\e{\varalpha}}$ is a $M\times M$ diagonal matrix with its elements having the values of the $\e\varalpha$ vector. To identify the terms contributing to the mean and variance, we have to reorder the parameters by applying the trace operator:

\begin{align*}
\sum_i\ej{\varc}{\varavec\vardvec\transpose\varf\transpose{\varp}\varp\varf\vardvec\transpose\varavec} =&  \trace{\sum_i\ej{\varc}{\varavec\vardvec\transpose\varf\transpose{\varp}\varp\varf\vardvec\transpose\varavec}}\\
=&  \trace{\sum_i\e{\transpose\varavec\varavec}\vardvec\e{\transpose\varf\transpose{\varp}\varp\varf}\vardvec}\\
=&  \sum_{mm\prime}(\sum_i\e{\transpose\varavec\varavec})_{mm\prime}(\vardvec\e{\transpose\varf\transpose{\varp}\varp\varf}\vardvec)_{mm\prime}\\
=&  \sum_{mm\prime}(\sum_i\e{\transpose\varavec\varavec})_{mm\prime}(\varc_{km}\e{\transpose\varf\transpose{\varp}\varp\varf}_{mm\prime}\varc_{km\prime})\\
=&  \varcvec\e{\transpose\vara\vara}\e{\transpose\varf\transpose{\varp}\varp\varf}\transpose\varcvec
\end{align*}

In a similar manner reordering the linear term gives the mean parameter:
\begin{align*}
\sum_i\e{\varsigmavec}\ej{\varc}{\varavec\vardvec\transpose{\varf}\transpose{\varp}\transpose{\varxvec}} =& \sum_i \e{\inverse{\varsigmavec}}\e{ \varavec}\vardvec\e{\transpose\varf}\e{\transpose{\varp}}\transpose{\varxvec}\\
=&\e{\varsigmavec}\sum_m\left(\varc_{km}\e{\transpose{\varf}_m}\e{\transpose{\varp}}\sum_i\transpose{\varxvec}\e{\varaele}\right)\\
=&\varcvec\e{\varsigmavec}\e{\transpose{\varf}}\e{\transpose{\varp}}\transpose{\varx}\e{\vara}
\end{align*}

By matching these expressions to the ones in \reff{sec:mmatch}, we get:

\begin{flalign}
	&\updateCsim&\\
	&\updateCmean&\\
	&\updateCvar 
\end{flalign}

\end{onecolumn}
\end{document}